\documentclass[runningheads]{llncs}

\usepackage{eccv}

\usepackage{eccvabbrv}
\usepackage{xr}
\usepackage{amsmath}
\usepackage{microtype}
\usepackage{graphicx}
\usepackage{url}
\usepackage{physics}
\usepackage{booktabs} 
\usepackage{multirow}
\usepackage{colortbl}
\usepackage{rotating}
\usepackage{enumitem}
\usepackage{lipsum}
\usepackage{subcaption}
\usepackage{xspace}
\usepackage{algorithm}
\usepackage{algpseudocode}
\usepackage{amssymb}
\usepackage{pifont}

\usepackage{makecell}
\usepackage{tcolorbox}
\usepackage{listings}

\usepackage[accsupp]{axessibility}  

\usepackage[pagebackref,breaklinks,colorlinks,citecolor=eccvblue]{hyperref}

\usepackage{orcidlink}

\begin{document}

\title{Sentinel: Embodied Cooperative Spatial\\Reasoning and Planning}



\author{
Xiangye Lin$^{1}$\thanks{denotes equal contribution.}
Hongxin Zhang$^{1*}$,
Ruxi Deng$^2$,
Qinhong Zhou$^1$,
Chuang Gan$^{1}$\\
$^1$ University of Massachusetts Amherst\\$^2$ University of Illinois Urbana-Champaign\\
\email{\{xiangyelin,hongxinzhang,qinhongzhou,chuangg\}@umass.edu}}
  
\authorrunning{Xiangye Lin, Hongxin Zhang, et al.}

\institute{}

\maketitle

\begin{abstract}
  In this work, we study \textit{Cooperative Spatial Intelligence}, the ability of decentralized embodied agents to coordinate effectively under dynamic environmental constraints across city-scale outdoor domains. We introduce \textit{Sentinel Challenge}, a benchmark where multiple decentralized embodied agents must communicate in natural language to agree on a mutually safe and convenient meeting point within large, city-scale outdoor environments. Each agent must then navigate safely while avoiding dynamic sentinels patrolling the area, with a tool providing coarse spatial information. To address this, we propose \textit{CoSaR} (Cooperative Spatial Reasoning and Planning), a framework that bridges the high-level communication and planning abilities of foundation models with the precision of classical spatial navigation algorithms. \textit{CoSaR} enables agents to exchange situational updates, reason over evolving spatial constraints, and collaboratively replan trajectories. Evaluated across 14 city-level scenes with 3–5 agents, \textit{CoSaR} consistently leads to faster gathering, shorter path lengths, and improved safety. Our results demonstrate that integrating dynamic communication with spatial reasoning is essential for robust multi-agent cooperation. By formalizing this new setting and providing a scalable benchmark, we aim to build a foundation for advancing cooperative spatial intelligence in embodied multi-agent systems. Code and challenge are available at \url{https://github.com/UMass-Embodied-AGI/Sentinel}.
\end{abstract}

\section{Introduction}
\label{sec:intro}

\begin{figure*}[t]
    \centering
    \includegraphics[width=1.0\linewidth]{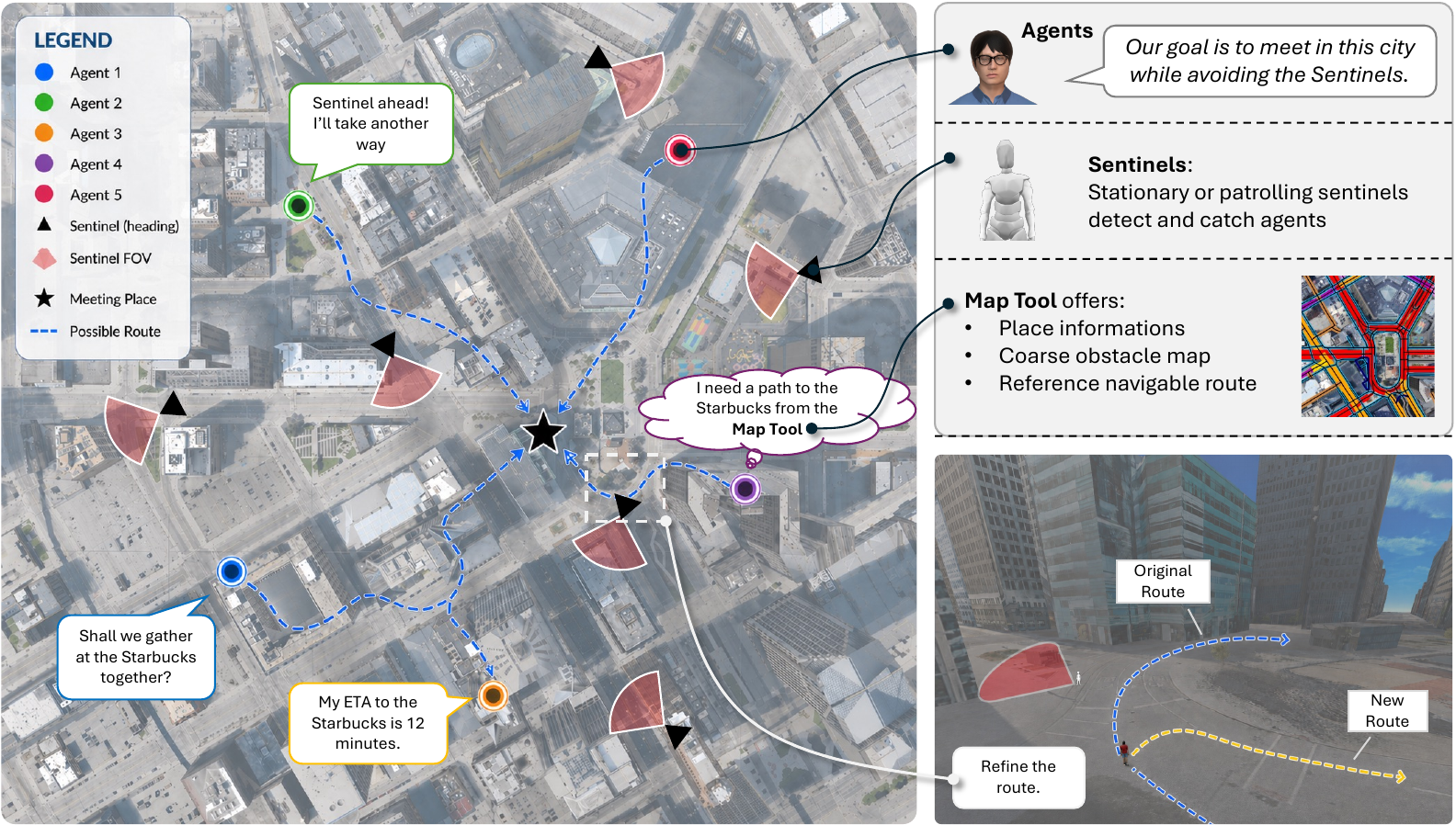}
    \caption{Embodied agents need to have \textbf{Cooperative Spatial Intelligence} to cooperate efficiently under dynamic spatial constraints in large spatial extents.}
    \label{fig:teaser}
\end{figure*}

Humans excel at coordinating in complex spatial environments, determining safe and convenient meeting points, sharing updates about obstacles, and adapting plans as situations evolve. We refer to this capability as \textit{Cooperative Spatial Intelligence}. Despite rapid progress in embodied AI, whether multi-agent systems can exhibit such competence in large, dynamic environments remains largely unexplored.
In this work, we study embodied multi-agent cooperation under dynamic spatial constraints in large-scale outdoor environments. As shown in Fig.~\ref{fig:teaser}, multiple decentralized embodied agents must communicate in natural language to agree on a mutually safe and convenient gathering point in large, city-scale outdoor scenes. After reaching consensus, each agent must navigate safely while avoiding dynamically moving sentinels patrolling the environment, relying on a map tool that provides only coarse point-to-point waypoints, an abstraction inspired by how humans coordinate while using a map to meet in a city.

Recent advances in foundation models~\cite{openai2023gpt4, guo2025deepseek} have significantly strengthened the capabilities of embodied multi-agent systems. Prior work, such as \textit{CoELA}~\cite{zhang2023building} demonstrates that agents equipped with large language models can communicate and cooperatively solve indoor rearrangement tasks, and \textit{CaPo}~\cite{liu2024capo} further improves cooperation efficiency through meta-plan generation. However, these systems operate only in confined, mostly static indoor environments where simple obstacle-avoidance navigation is sufficient. Scaling cooperation to realistic outdoor settings introduces new challenges: vastly larger spatial extents, path uncertainties, dynamically changing constraints, and the need for continual multi-agent information exchange.

To address these challenges, we introduce \textbf{CoSaR}, a cooperative spatial reasoning and planning framework that combines the communication and planning strengths of foundation models with classical spatial navigation algorithms. CoSaR maintains a structured spatial memory consisting of a low-level occupancy map and high-level representations of each agent and patrolling sentinels. This memory is continually updated using visual observations and natural-language communications, enabling agents to infer missing information, query a map tool or peers to gather additional context, refine coarse waypoints, and collaboratively replan routes in real time as conditions evolve.

We instantiate this problem setting within Virtual Community~\cite{zhou2025virtual} and develop the \textbf{Sentinel} Challenge featuring configurable difficulty based on sentinel numbers and moving patterns. Experiments across 14 large-scale city environments with 3–5 agents demonstrate that CoSaR achieves consistently faster gathering, shorter path lengths, and substantially higher safety compared to strong baselines such as CoELA~\cite{zhang2023building} and RoCo~\cite{mandi2023roco}. These improvements stem from CoSaR’s ability to integrate dynamic communication, structured spatial reasoning, and continual adaptation to changing environmental constraints. By formalizing this new problem space and releasing a scalable benchmark, we aim to advance the study of cooperative spatial intelligence in embodied multi-agent systems through the Sentinel Challenge and CoSaR methods.

Our contributions:

\begin{itemize}
    \item We introduce the \textbf{Sentinel} Challenge to evaluate the embodied agent's \textit{Cooperative Spatial Intelligence}, the capability to cooperate effectively under dynamic spatial constraints.
    \item We propose \textbf{CoSaR}, an embodied cooperative spatial reasoning and planning method featuring a dynamic spatial memory that enables agents to share information, reason about spatial dynamics, and replan collaboratively.
    \item Experiments across 14 city-scale scenes with 3–5 agents show that our method effectively leverages spatial information and dynamic communication, outperforming strong baselines across multiple metrics.
\end{itemize}
\section{Related Work}

\subsection{Multi-Agent Cooperation}

Multi-agent cooperation is a long-standing and fundamental research problem in artificial intelligence~\cite{stone2000multiagent, lowe2017multi,samvelyan2019starcraft,carroll2019utility, suarez2019neural, jaderberg2019human, amato2019modeling, Baker2020Emergent,bard2020hanabi, jain2020cordial, wen2022multi, szot2023adaptive}. For embodied intelligence, existing works primarily focus on multi-agent cooperation in household environments~\cite{puigwatch, puig2023habitat}, while recent studies further extend this setting to communicative scenarios~\cite{zhang2023building, liu2024capo, zu2025collaborative, zhang2025ella}. Different from the existing works, in this paper, we focus on the multi-agent cooperation problem in dynamic city-level environments, where agents are often spatially distant and rarely encounter each other directly, requiring effective communication strategies, complex spatial reasoning, and robust inference over continuously changing environments.

\subsection{Foundation Models for Embodied AI}
The emerging foundation models have shown significant success in building autonomous agents, 
especially with recent advances in both general and embodied settings~\cite{wang2023survey, xi2023rise, sumers2023cognitive, puig2023habitat, savva2019habitat, li2023behavior, padmakumar2022teach, kolve2017ai2, misra2018mapping, xia2018gibson, xiang2020sapien, zhou2024hazard}. 
Specifically, large language models are widely used for planning and decision-making with text input~\cite{wang2023describe, ahn2022can, sharma2021skill, wang2023voyager, park2023generative}, 
while vision-language models are adopted to facilitate end-to-end planning from visual perceptions~\cite{zitkovich2023rt, jiang2023vima, wang2023jarvis, zhang2024combo, kim2024openvla, mon2025embodied}. 
In this paper, we use foundation models as the backbones of our agent framework for perception, communication, and decision-making.

\subsection{Embodied Navigation and Spatial Intelligence}
Embodied navigation~\cite{anderson2018evaluation, wijmans2019dd, shah2021ving, gadre2022clip, majumdar2022zson, shah2023lm, huang2022visual, khandelwal2022simple, song2023llm, zheng2024towards} focuses on how agents perceive, reason, and act within 3D environments to reach goals based on visual and linguistic inputs. The problem studied in this paper is closely related to the subfield of embodied urban navigation~\cite{morales2009autonomous, kummerle2013navigation, kahn2021badgr, shah2023vint, sridhar2024nomad, liu2025citywalker}, which extends navigation into outdoor and city-scale environments, emphasizing long-horizon planning, multimodal perception, and real-world constraints. A key capability for tackling such problems is spatial reasoning~\cite{hong20233d, wang2024embodiedscan, fu2024scene, yang2025thinking, yang20253d, cai2025spatialbot}, which enables agents to infer geometric structures, spatial relations, and affordances from limited observations to support effective decision-making in complex environments. In this paper, we introduce a novel spatial reasoning framework to address these challenges.
\section{Problem Statement}
\label{sec:problem}

We study the problem of embodied multi-agent cooperation under dynamic spatial constraints. A group of decentralized embodied agents is randomly initialized across a large, city-scale outdoor environment. Their objective is to communicate in natural language and navigate safely to gather at a \textbf{Place} as soon as possible. The environment contains dynamically changing hazards—modeled as \textbf{Sentinels}—that can detect and ``catch'' agents within their ego-centric field of view. The task succeeds \emph{only if all agents safely gather at a place} without being caught.
To support navigation across vast spatial scales, agents may query a \textbf{Map Tool} that provides only coarse, high-level spatial information (e.g., waypoint routes), mirroring how humans coordinate using a city map when meeting in an unfamiliar area.

More formally, our problem can be formalized as a decentralized partially observable Markov decision process with communication (DEC-POMDP-COM)~\cite{bernstein2002complexity, spaan2006decentralized, zhang2023building}, defined by $(n, S, {A_i}, {O_i}, T, G, h)$, where 
\begin{itemize}
    \item $n$ is the number of agents.
    \item $S$ is a finite set of world states.
    \item $A_i = A^{W} \cup A^{C} \cup A^{M}$ is the action space for agent $i$, consisting of
    \begin{itemize}
        \item $A^{W}$ actions for interacting with the world, such as navigation and task-completion signals,
        \item $A^{C}$ communication actions,
        \item $A^{M}$ map tool query actions.
    \end{itemize}
    \item $O_i = O^{V} \times O^{C} \times O^{M}$ is the observation space for agent $i$, including
    \begin{itemize}
        \item $O^{V}$ visual observations,
        \item $O^{C}$ messages received from other agents,
        \item $O^{M}$ results returned by the map tool.
    \end{itemize}
    \item $T$ represents the joint transition model.
    \item $G$ represents the task goal.
    \item $h$ denotes the task horizon.
\end{itemize}

\textbf{Places} are predefined indoor or outdoor landmarks in the environment that may serve as candidate gathering points. Agents are also initialized at indoor places to prevent them from being immediately caught by sentinels. Each city scene contains 50–150 Places. Initially, each agent knows a part of these places. To get information about unknown places, they must query the Map Tool.

\textbf{Sentinels} represent dynamic spatial hazards that constrain safe navigation. We include two types:

\begin{itemize}
    \item \emph{Stationary Sentinels}: remain fixed and rotate continually.
    \item \emph{Patrolling Sentinels}: move along predefined patrol routes.
\end{itemize}

Both types of sentinels stay outdoors and cannot detect entities indoors.
A sentinel initiates a warning countdown when an agent stays within its ego-view. When starting the countdown, it will send a \emph{warning signal} to the agent. The countdown decreases faster when the agent is closer. An agent is considered \emph{caught} when the countdown reaches zero, and will be transported outside of the city immediately. More formally,

$$\text{If } \sum_{p \in G}[label_p = label_{agent}]>\frac{|G|}{1000}, \text{ countdown starts}$$

Where $G$ is the RGB-D image, $label_p$ is the label of pixel $p$ based on the segmentation result, and $label_{agent}$ is the label of this agent. Countdown time $t$ is initially set to $15$ when it starts. Then every second, 
\[
t=t-1000\times \frac{\sum_{p \in G}[label_p = label_{agent}]}{|G|}
\]
If the agent exits the sentinel's view, the countdown resets, allowing escape but requiring coordinated and cautious path planning.

\textbf{Map Tool} provides coarse spatial information through five query types:

\begin{itemize}
    \item \textit{QueryRoute$(p_1, p_2)$}: returns a waypoint-based route between coordinates $p_1$ and $p_2$.
    \item \textit{QueryNearby$(p_1)$}: returns a list of places near coordinate $p_1$.
    \item \textit{QueryPlace$(place)$}: returns metadata for a given place $(place)$, including its location and bounding box.
    \item \textit{QueryMap$()$}: returns an aerial-view image of the whole scene. The map is represented as a binary image, where white denotes navigable space. It is derived from the Virtual Community dataset and serves only as a coarse approximation of the environment rather than an exact representation of the physical layout.
    \item \textit{QueryRefinedRoute$(list[p])$}: return a waypoint-based route. The Map Tool first maps each input coordinate $p$ to the nearest valid waypoint, and then connects the waypoints sequentially, using the same algorithm as \textit{QueryRoute}.
\end{itemize}

The Map Tool is implemented based on the Virtual Community dataset. It generates waypoints in the city environment, constructs a connected waypoint graph, and performs \textit{QueryRoute} operations using the Dijkstra algorithm~\cite{dijkstra1959note}. Additional implementation details are provided in the supplementary materials.

The tool intentionally exposes only coarse, high-level spatial information, requiring agents to communicate, reason about incomplete spatial knowledge, and collaboratively refine plans under uncertainty—mirroring real-world human coordination in large-scale environments.
\begin{figure*}[t]
    \centering
    \includegraphics[width=1.0\linewidth]{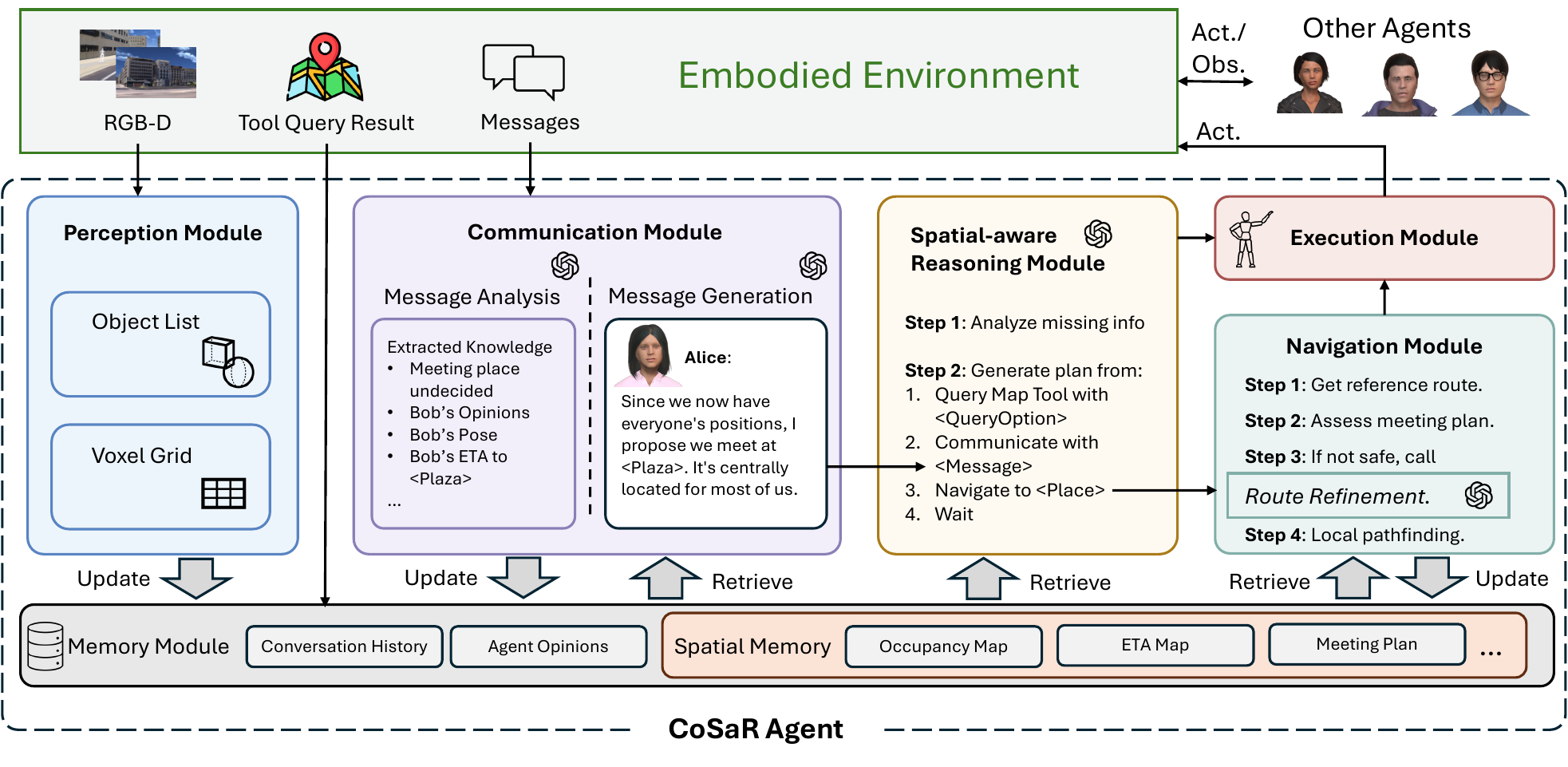}
    \caption{\textbf{Method overview.} CoSaR maintains a Spatial Memory that integrates perceived visual observations, map-tool responses, and spatial information extracted from natural-language communication. A spatial-aware reasoning module then uses this memory to plan the actions.}
    \label{fig:method}
    \vspace{-4mm}
\end{figure*}

\section{CoSaR: Embodied Cooperative Spatial Reasoning and Planning}

CoSaR equips embodied agents with a spatially grounded reasoning architecture that integrates perception, memory, and planning. After receiving egocentric sensory input and natural-language messages from teammates, the agent first processes visual signals through a dedicated \textit{Perception Module}, which extracts objects, identifies sentinels, and produces semantic segmentation suitable for spatial reasoning (section~\ref{sec:perception}). The natural-language messages are analyzed by the \textit{Communication Module}, extracting useful spatial information and cooperation information (section~\ref{sec:comm}). These heterogeneous inputs are converted into structured spatial information such as updated poses, travel-time estimates, occupancy patterns, and sentinel danger zones and consolidated into a persistent \textit{Spatial Memory} (Section~\ref{sec:memory}), which stores the agent’s evolving understanding of the environment, including meeting plans, occpupancy maps encoding traversability and known poses of teammates and sentinels, estimated arrival-time relationships. The memory serves as a unified state representation supporting both communication and navigation. Building on this memory, the \textit{Spatial-aware Reasoning Module} plans the next action by deciding when to communicate with others, query the map tool, or wait, or by guiding movement through the environment (section~\ref{sec:reasoner}). When the agent identifies that its current route intersects a dangerous region, it invokes the \textit{Route Refinement} module, which reconstructs a schematic map from spatial memory and solicits spatial reasoning from a VLM to propose a safer route before obtaining an actionable path from the map tool (section~\ref{sec:navigation}). Together, these components enable agents to perceive, communicate, and navigate in a spatially grounded manner, forming the core of CoSaR’s embodied cooperative spatial reasoning and planning capabilities.

\subsection{Perception Module}
\label{sec:perception}
The agent first processes the observed raw RGB-D input with open-set object detection models~\cite{liu2023grounding} to identify all visible objects $o$, including other agents, and then uses Segment Anything~\cite{ravi2024sam2segmentimages} to predict segmentation masks $m$, and projects each object into a point cloud $p$ with the depth input. To supplement with semantic features, the agent calculates the CLIP feature~\cite{radford2021learning} for each object with cropped rgb images with respective segmentation masks. A voxel grid $g$ is also constructed with the entire RGB image and depth map for later use. Specifically, the agent matches each perceived object with the sentinel features to identify the existence of the sentinels, and marks the grids occupied by sentinels with warning labels. Therefore, the RGB-D observation is perceived into a list of objects $o(p, f)$ and a voxel grid $g$.

\subsection{Communication Module}
\label{sec:comm}
The agent uses the communication module to analyze messages received from other agents and generate potential messages to exchange based on current knowledge of the environment and other agents.

If there are communication messages, the agent uses LLM to extract potentially useful information, including spatial information such as the current positions of other agents and sentinels, ETA information from an agent to a specific place, and cooperation information such as the opinions of each teammate. With this updated information, the agent determines the most reasonable target location for gathering.

To facilitate planning, CoSaR generates potential communication messages to send at each step. To generate reasonable communication messages that facilitate the task, the communication module utilizes LLMs as a message generator and prompts it with key components of our spatial memory, including \textit{Poses list}, \emph{ETA map}, \emph{available places}. We also equipped it with the stored conversation history and previously analyzed opinions of each teammate.

\subsection{Spatial Memory}
\label{sec:memory}

The spatial memory is the core component of our agent that stores the key knowledge about the environment and other teammates. The basic components of the spatial memory include:

\noindent\textbf{Occupancy map and danger zones} The agent constructs a local occupancy map that captures ground accessibility with the perceived voxel grids. The known locations of all agents and sentinels are also noted on the map. To support the meeting challenge, the agent further designates \textbf{danger zones} surrounding all known sentinel locations. This map is maintained through a combination of visual perception and information exchange from communication.

\noindent\textbf{Estimated Time of Arrival (ETA) Map} The ETA map records the estimated travel time of each agent to each candidate place. The agent infers others’ ETAs through communication and obtains its own by querying the map tool.

\noindent\textbf{Meeting plan} The meeting plan consists of a selected meeting place and a reference route to it. While the meeting place is continuously negotiated and updated through communication, the reference route is initially obtained from the map tool via the \textit{Query Route} action and subsequently maintained through iterative refinement during navigation.

\subsection{Spatial-aware Reasoning Module}
\label{sec:reasoner}

The Spatial-aware Reasoning Module serves as the central cognitive engine, synthesizing high-level coordination strategies from the agent's underlying spatial memory. This module systematically interprets stored spatial trajectories and environmental constraints to orchestrate communication and navigation in a decentralized manner.

The spatial-aware reasoning module scans the spatial memory to determine what information remains missing, what should be queried, and what actions are most beneficial for progressing the discussion. We leverage the zero-shot reasoning capabilities of a Large Language Model (LLM), prompting it with a structured representation of the spatial memory. This includes the Pose Registry (agent coordinates), an Estimated Time of Arrival (ETA) Map, and a list of Candidate Meeting Places. By integrating these with the historical dialogue and the inferred preferences of teammates, the LLM selects from four plan types:
\begin{itemize}
    \item \textbf{Query}: Invokes the map tool with specific parameters to reduce spatial uncertainty.
\item \textbf{Communicate}: Broadcasts a natural language message synthesized by the communication module to align teammate beliefs.
\item \textbf{Navigate}: Initiates a trajectory toward a prioritized spatial goal.
\item \textbf{Wait}: Temporarily halts movement to synchronize with teammate arrivals or await critical information.
\end{itemize}

Each selected action is accompanied by a justification, providing a trace of the agent's reasoning process. Navigation-type plans are subsequently offloaded to the Navigation Module (Sec. \ref{sec:navigation}) for low-level execution and obstacle avoidance.

To reduce inference cost, the spatial-aware reasoning module is activated only under the following conditions:
\begin{itemize}
\item A new event is observed (e.g., incoming messages or newly detected sentinels).
\item The communication module detects disagreement on the meeting place.
\item At regular intervals (every 120 seconds).
\end{itemize}

\begin{figure*}[t]
    \centering
    \includegraphics[width=0.95\linewidth]{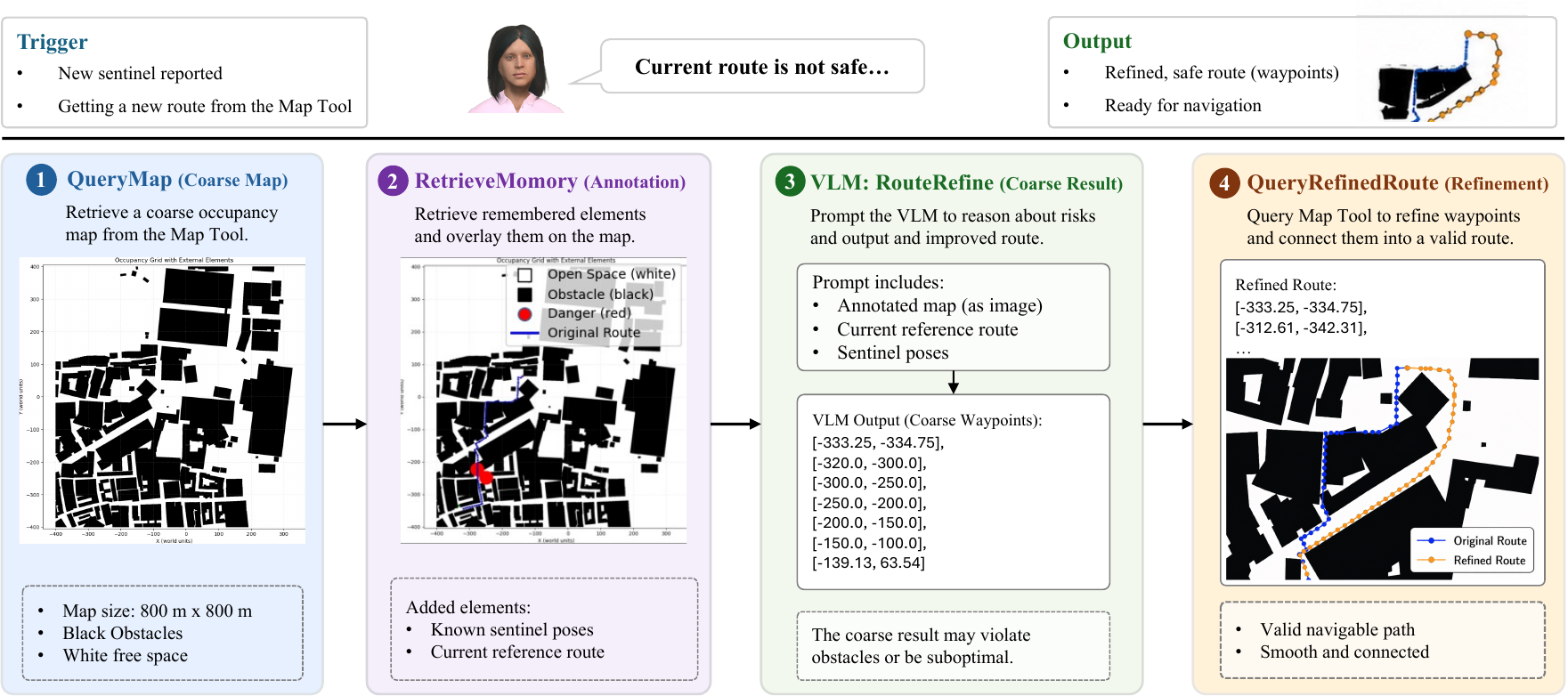}
    \caption{\textbf{Route Refinement}. The agent will call \textit{QueryMap} and get a coarse-level 2D-map image, which is dynamically augmented with visual prompts representing the current route (blue) and danger zones (red) corresponding to sentinel locations. A Vision-Language Model (VLM) performs zero-shot spatial reasoning over this annotated visual input to synthesize a new route that ensures safe navigation toward the meeting point while maintaining a safe buffer from dynamic threats.}
    \label{fig:refine}
    \vspace{-8mm}
\end{figure*}

\subsection{Navigation Module}
\label{sec:navigation}

The navigation module translates high-level navigation plans into executable dynamic hazard-aware trajectories. Given a target destination, the agent first queries the map tool for a baseline global route. This route is subjected to a Safety Assessment phase, where it is cross-referenced against known Danger Zones. If a safety violation is detected, the agent triggers a multi-stage Route Refinement process to synthesize a viable detour.

\noindent\textbf{Route Refinement via Multimodal Grounding.} When a route is flagged as unsafe, the agent invokes a \textit{QueryMap} action to retrieve a coarse-scale, top-down aerial representation of the environment. As illustrated in Fig. \ref{fig:refine}, we construct a Schematic Spatial Map by augmenting this visual representation with metadata from the agent's spatial memory, including current ego-pose, detected sentinel coordinates, and the current reference path.
This augmented map is then passed to a Vision-Language Model (VLM). We provide the VLM with explicit coordinates for the agent, the destination, and the intersection points of the danger zones and the reference route. The VLM is prompted to return a series of refined coordinates. To account for the inherent spatial stochasticity and potential hallucinations of the VLM, the model’s proposed trajectory is passed through a \textit{QueryRefinedRoute} verification step. This step maps the VLM's high-level route back to the valid navigation waypoints that the agent can directly follow. To ensure robustness against repeated VLM failures, we implement a retry-limit mechanism, falling back to a heuristically-modified map-tool route, where we simply remove the waypoints intersecting with any danger zone, if a valid path is not synthesized within $N$ attempts.

\noindent\textbf{Emergency Avoidance} During navigation, certain circumstances may rise, such as a sentinel emerging suddenly after a corner. If the agent decides it is in an immediately dangerous pose, it will perform an emergency avoidance. The further details are included in the \textbf{Appendix}.

\noindent\textbf{Low-level Control.}
Once the refined waypoints are finalized, the agent employs the $A^*$ algorithm for local pathfinding based on the occupancy map, ensuring fine-grained obstacle avoidance and adherence to the global trajectory in a continuous action space.

\section{Sentinel: An Embodied Cooperative Spatial Reasoning and Planning Benchmark}
\label{sec:benchmark}

We instantiate the above problem in a city-scale environment using the \textit{Virtual Community} platform~\cite{zhou2025virtual}.

\paragraph{Environment.}
The benchmark consists of 24 fully annotated scenes. Each scene spans approximately $800\,\mathrm{m}\times800\,\mathrm{m}$ and is constructed from real-world 3D reconstructions. Agents receive egocentric $512\times512$ RGB-D observations and interact with the environment through navigation, communication, and map queries.

\paragraph{Places.}
Each scene contains 50--150 predefined places, which serve as candidate gathering points. Agents initially possess only partial knowledge of these places and must discover additional locations through interaction with the environment and the Map Tool. 

\paragraph{Agents.}
Each scene includes up to 15 agents with predefined configurations, including their names, initial places, and knowledge. The number of participating agents in each run is controlled by configurable parameters.

\paragraph{Map Tool Implementation.}
The Map Tool is constructed from OpenStreetMap data by extracting road networks, generating waypoints, and building a connected topological graph. Route queries are computed using the Dijkstra algorithm~\cite{dijkstra1959note}. The tool intentionally provides only coarse, high-level spatial information. Additional implementation details are provided in the appendix.

\paragraph{Sentinels.}
Each scene includes up to 20 sentinels (both stationary and patrolling) with predefined configurations. Sentinels are initialized at random locations within the scene. To prevent trivial solutions, a subset of sentinels (e.g., the first five) is placed near the geometric center of agents’ initial places, increasing task difficulty. The number of active sentinels in each run is also controlled via configurable parameters.

\paragraph{Metrics.}
The benchmark reports the following metrics:
\begin{itemize}[leftmargin=1.5em]
    \item \textbf{Success Rate}: fraction of episodes in which all agents successfully gather within a horizon of 1500 steps.
    \item \textbf{Caught Rate}: fraction of agents that are captured by sentinels.
    \item \textbf{Detected Rate}: proportion of time steps in which any agent is detected by a sentinel.
    \item \textbf{Time Cost}: number of steps taken for all agents to gather.
    \item \textbf{Distance Traveled}: aggregate distance traveled by all agents.
\end{itemize}

\section{Experiment}

\subsection{Experimental Setup}

\paragraph{Sentinel Challenge.} We evaluate our method in the Sentinel Benchmark~\ref{sec:benchmark}. Task difficulty is controlled by varying the number of sentinels and their motion patterns. We conducted experiments with 3 agents and 5 sentinels of two different moving patterns as introduced in section~\ref{sec:problem}, over 14 diverse scenes. 

\paragraph{Sentinel Challenge w/ Oracle Perception.} We also evaluate a setting with oracle perception, in which ground-truth segmentation is provided during perception. This significantly improves the probability of correctly detecting a sentinel.

\paragraph{Baselines} We compare against six representative baselines:

\begin{itemize}
    \item \textbf{Oracle Centered}: selects the meeting location as the geometric center of all agents, assuming oracle access to their ground-truth positions.
    \item \textbf{Oracle Centered w/ DZ}: extends the above with danger-zone–based sentinel avoidance (Section~\ref{sec:memory}).
    \item \textbf{MCTS Planner}: selecting the meeting location using MCTS planners. Each agent uses an MCTS planner to minimize the distance to other agents, the total traveled distance, and the detection rate.
    \item \textbf{RoCo}~\cite{mandi2023roco}: a strong multi-agent cooperation method adapted to our setting. RoCo represents methods that plan first and execute later: agents communicate only at the beginning of the task to agree on a meeting place and the full navigation plan. 
    \item \textbf{CoELA}~\cite{zhang2023building}: a modular framework powered by foundation models with dynamic communication and planning, enhanced with danger-zone avoidance.
    \item \textbf{MAT}~\cite{wen2022multi}: Multi-Agent Transformer adapted to our setting following the CoELA paper. A shared encoder–decoder Transformer consumes a top-down semantic map plus per-agent features and autoregressively emits joint meeting-place decisions over the same top-$K$ candidate set, trained online with PPO. MAT serves as a centralized end-to-end multi-agent RL baseline.
\end{itemize}

\paragraph{Implementation Details} We use \texttt{gpt-4o} as the foundation model backbones for all methods unless otherwise specified. Experiments are conducted on A16 GPUs.

\subsection{Results}

\begin{table*}[t]
\centering
\caption{\textbf{Results for 5 agents and 10 sentinels} We report the average score over 14 scenes and 6 runs here. Best performance is shown in \textbf{bold}. The Oracle Centered baseline is excluded from the Oracle-perception scenario, since the method itself does not utilize Oracle perception.}
\label{tab:main_5_10}
\resizebox{\linewidth}{!}{
\begin{tabular}{l|ccccc|ccccc}
\toprule
& \multicolumn{5}{c|}{\textbf{10 Stationary Sentinels}} 
& \multicolumn{5}{c}{\textbf{10 Patrolling Sentinels}} \\
\cmidrule(r){2-6} \cmidrule(l){7-11}
Method &
\makecell{Succ.\\Rate$\uparrow$} &
\makecell{Caught\\Rate$\downarrow$} &
\makecell{Detect.\\Rate$\downarrow$} &
Time$\downarrow$ &
Dist.$\downarrow$
&
\makecell{Succ.\\Rate$\uparrow$} &
\makecell{Caught\\Rate$\downarrow$} &
\makecell{Detect.\\Rate$\downarrow$} &
Time$\downarrow$ &
Dist.$\downarrow$
\\
\midrule
\multicolumn{11}{c}{\textit{\textbf{Sentinel Challenge}}} \\
\midrule
Oracle Centered
& 7.14 & 51.43 & 1.80 & 1429.86 & 1800.07
& 14.29 & 28.57 & 1.50 & 1367.07 & 1401.93
 \\
Oracle Centered w/ DZ
& 11.90 & 39.29 & 1.86 & 1401.40 & 1598.50
& 29.76 & 22.14 & 1.66 & \textbf{1228.89} & 1381.91
 \\
MCTS
& 10.71 & 47.38 & 2.21 & 1371.26 & 2206.20
& 15.48 & 40.48 & 2.02 & 1338.67 & 2226.39
 \\
RoCo
& 26.19 & 26.90 & 1.13 & 1255.81 & \textbf{1333.27}
& 29.76 & 28.57 & 1.73 & 1306.00 & 1480.76
 \\
CoELA
& 16.67 & 25.24 & \textbf{1.08} & 1340.77 & 1522.53
& 26.19 & \textbf{17.62} & 0.94 & 1304.94 & 1436.53
 \\
MAT
& 0.00 & 35.71 & 1.13 & 1500.00 & 1453.82
& 7.14 & 20.24 & \textbf{0.78} & 1447.17 & \textbf{1269.10}
 \\
CoSaR (Ours)
& \textbf{32.14} & \textbf{21.43} & 1.27 & \textbf{1194.45} & 1552.45
& \textbf{32.14} & 27.62 & 2.22 & 1246.86 & 1577.99
 \\
\midrule
\multicolumn{11}{c}{\textit{\textbf{Sentinel Challenge w/ Oracle Perception}}} \\
\midrule
Oracle Centered w/ DZ
& 28.57 & 24.29 & 2.09 & 1237.50 & 1459.10
& 35.71 & 19.76 & 0.95 & \textbf{1230.06} & 1393.05
 \\
MCTS
& 33.33 & 22.38 & 2.55 & 1196.80 & 2147.75
& 19.05 & 31.43 & 2.26 & 1309.76 & 2393.93
 \\
RoCo
& 46.43 & 15.00 & 1.44 & 1106.30 & 1332.00
& \textbf{38.10} & 17.14 & 1.41 & 1268.10 & 1472.20
 \\
CoELA
& 34.52 & 9.52 & 1.24 & 1188.02 & \textbf{1237.82}
& 36.90 & \textbf{10.95} & 0.66 & 1248.24 & 1350.23
 \\
MAT
& 7.14 & 19.76 & 1.00 & 1455.43 & 1815.91
& 0.00 & 11.43 & \textbf{0.49} & 1500.00 & \textbf{1315.29}
 \\
CoSaR (Ours)
& \textbf{53.57} & \textbf{9.05} & \textbf{0.97} & \textbf{1080.46} & 1387.31
& \textbf{38.10} & 20.95 & 1.72 & 1247.62 & 1708.91
\\
\bottomrule
\end{tabular}
}
\end{table*}

\begin{figure*}[t]
    \centering
    \includegraphics[width=1.0\linewidth]{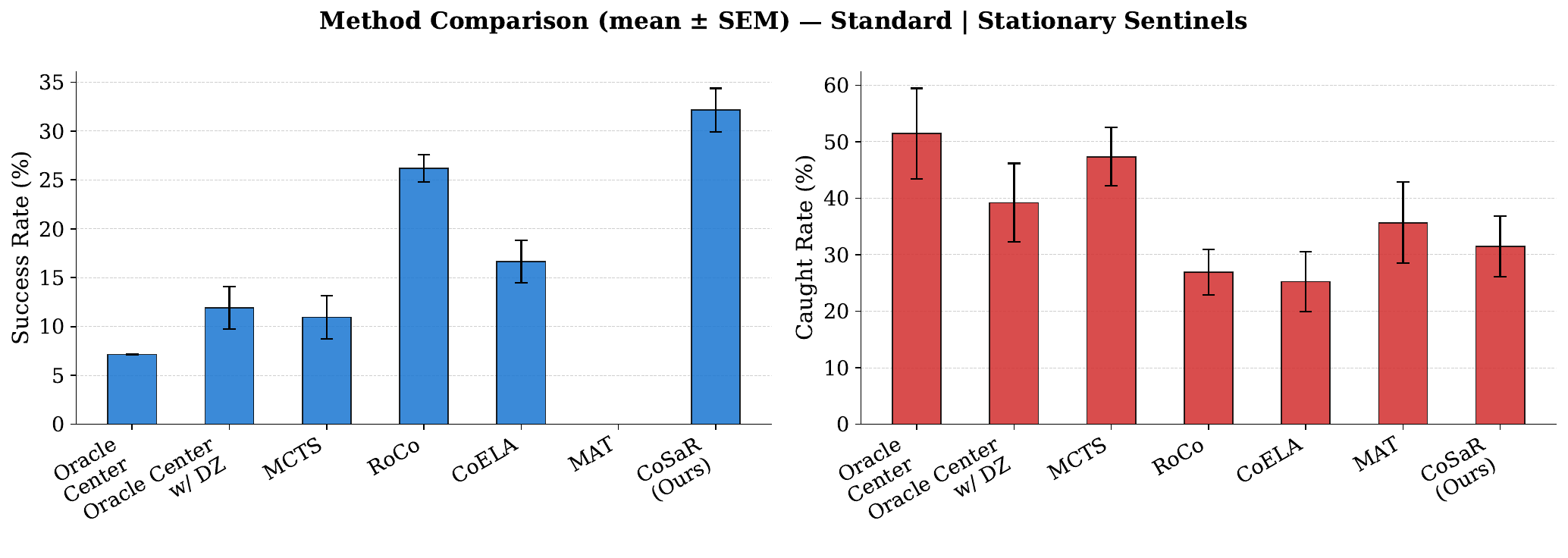}
    \caption{Method comparison under the standard setting with stationary sentinels. Results are reported as mean ± SEM over 14 scenes with 5 agents and 10 sentinels (6 runs per scene). Left: task success rate (\%). Right: caught rate (\%).}
    \label{fig:error_bar}
\end{figure*}

\begin{figure*}[t]
    \centering
    \includegraphics[width=1.0\linewidth]{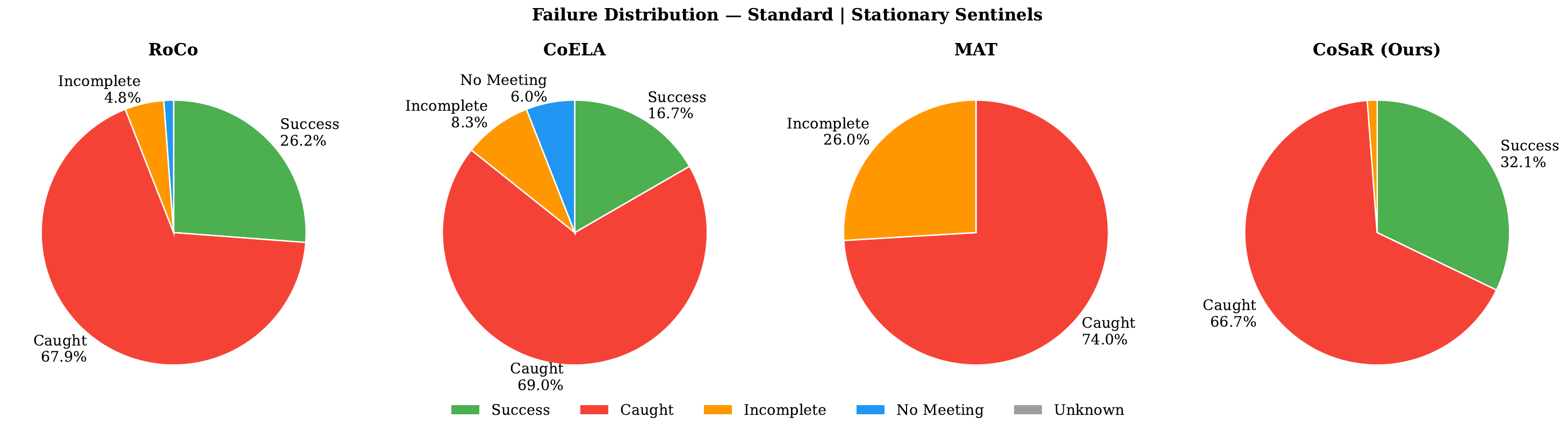}
    \caption{\textbf{Failure breakdown (\%)} under the setting of 5 agents and 10 stationary sentinels over 14 scenes and 6 runs. Most failures lie in unsafe decisions (\textbf{Caught}).}
    \label{fig:failure}
\end{figure*}

\paragraph{Failure Reasons.} We show the failure reasons breakdown in Figure~\ref{fig:failure}. The reasons for task failure are threefold. First, some agents are captured by sentinels. Second, some agents take too long to reach the target location, exceeding the time limit. Third, agents may issue a task completion signal while not being located at a common meeting place.

\paragraph{CoSaR Agents generally achieve a higher success rate with lower cost.} As shown in Table~\ref{tab:main_5_10}, CoSaR agents consistently obtain the highest success rate across tasks. The high performance of RoCo is also predictable, as we incorporate part of the spatial memory into its implementation. CoSaR also has the overall lowest \textit{Time} and \textit{Distance} cost, indicating more efficient coordination and faster convergence.

\paragraph{Spatial memory is important.} Without dynamic discussion, Oracle Centered Agents and RoCo agents may get stuck at unsafe meeting points. However, their performance remains considerably higher than that of CoELA agents, which support dynamic communication but lack any spatially aware module. This contrast highlights the essential role of spatial memory in effective coordination.

\begin{figure*}[t]
    \centering
    \includegraphics[width=1.0\linewidth]{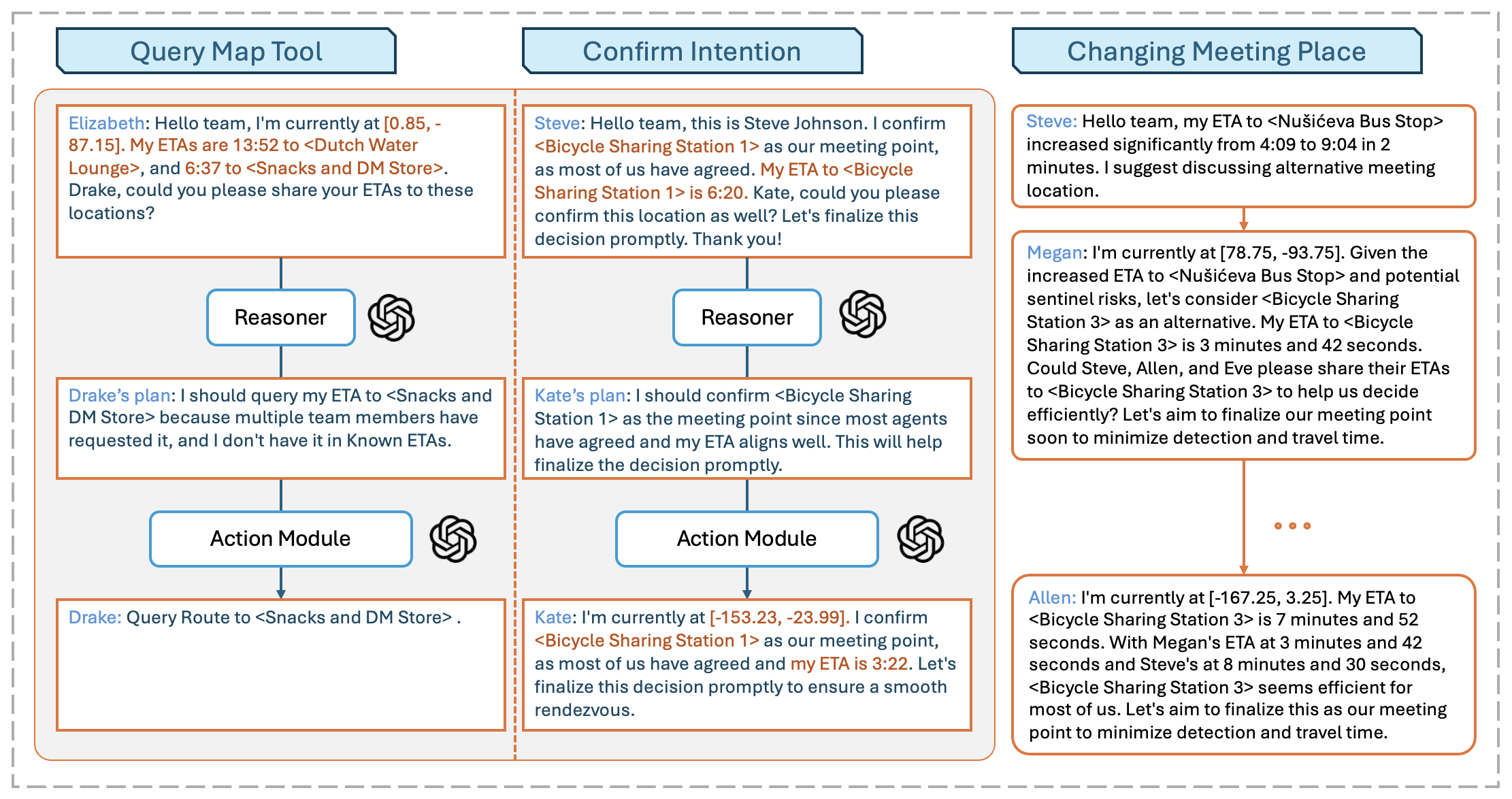}
    \caption{\textbf{Case Study.} The left panels demonstrate how CoSaR agents decide their next action when actively communicating. The right panel showcases an example of how CoSaR agents successfully change their meeting place upon changing circumstances. More case studies are provided in the \textbf{Appendix}.}
    \label{fig:qualitative}
\end{figure*}

\paragraph{CoSaR Agents can communicate effectively to exchange spatial information.} Figure~\ref{fig:qualitative} illustrates the qualitative communication process among the agents. The examples demonstrate that, supported by our framework’s reasoning and planning capabilities, the agents can respond appropriately to communicative requirements and pose pertinent questions that help advance the discussion. During the meeting procedure, certain circumstances may arise, like a new sentinel or invalid reference route, which will result in an increase in the current estimated time to arrival. Through coordinated decision-making, the agents are also able to address unexpected difficulties that arise during the meeting process, adjusting their meeting location accordingly.

\begin{table}[t]
\centering
\small

\caption{\textbf{CoSaR performs best across different numbers of agents and sentinels.}
We report the mean success rate and detected rate under the patrolling sentinel setting over 14 scenes and 2 runs.}
\label{tab:agents_num}
\resizebox{\linewidth}{!}{
\begin{tabular}{l|cc|cc|cc|cc}
\toprule
\multirow{2}{*}{Method} 
& \multicolumn{2}{c|}{\makecell{\textbf{3 Agents} \\ \textbf{10 sentinels}}} 
& \multicolumn{2}{c|}{\makecell{\textbf{5 Agents} \\ \textbf{5 sentinels}}}  
& \multicolumn{2}{c|}{\makecell{\textbf{5 Agents} \\ \textbf{10 sentinels}}}  
& \multicolumn{2}{c}{\makecell{\textbf{5 Agents} \\ \textbf{20 sentinels}}} \\
\cmidrule(r){2-3} \cmidrule(r){4-5}
\cmidrule(r){6-7} \cmidrule(l){8-9}
& \makecell{Success\\Rate\,$\uparrow$} 
& \makecell{Caught\\Rate\,$\downarrow$} 
& \makecell{Success\\Rate\,$\uparrow$} 
& \makecell{Caught\\Rate\,$\downarrow$} 
& \makecell{Success\\Rate\,$\uparrow$} 
& \makecell{Caught\\Rate\,$\downarrow$} 
& \makecell{Success\\Rate\,$\uparrow$} 
& \makecell{Caught\\Rate\,$\downarrow$} \\
\midrule
Oracle Center                   & 21.43 & 40.48 & 7.14 & 51.43 & 7.14 & 51.43 & 7.14 & 57.14 \\
Oracle Center w/ DZ            & 21.43 & 40.48 & 14.29 & 32.86 & 7.14 & 38.46 & 7.17 & 44.62 \\
MCTS                            & 25.00 & 33.33 & 17.86 & 40.00 & 7.14 & 50.71 & 7.14 & 40.77 \\
RoCo~\cite{mandi2023roco}      & 28.57 & 27.38 & 35.71 & 25.71 & 28.57 & 24.29& 17.86 & 36.43 \\
CoELA~\cite{zhang2023building} & 39.29 & \textbf{26.19} & 21.43 & 22.14 & 21.43 & \textbf{21.43} & 7.14 & 43.08 \\
CoSaR (Ours)                    & \textbf{42.86} & 27.38 & \textbf{39.29} & \textbf{20.00} & \textbf{35.71} & 24.29 & \textbf{21.43} & \textbf{32.14} \\
\bottomrule
\end{tabular}
}
\end{table}

\paragraph{CoSaR performs best across different numbers of agents and sentinels.} To evaluate whether our proposed method is effective across different numbers of agents and sentinels, we experimented with three additional settings:
\begin{itemize}
    \item 3 agents, 10 sentinels.
    \item 5 agents, 5 sentinels.
    \item 5 agents, 20 sentinels.
\end{itemize}
The results are shown in Table~\ref{tab:agents_num}. As shown in the results, increasing the agent number or the sentinel number makes the challenge overall harder to complete, since the possibility of an agent being caught gets higher, and it requires more agents to arrive at the same place. CoSaR maintains a consistently lower caught rate across varying numbers of agents, with its advantage increasing as the team size grows. Moreover, CoSaR prevents its success rate from deteriorating as sharply as the other methods, demonstrating stronger robustness under multi-agent settings.

\paragraph{Efficiency-Safety Tradeoff}

As shown in Table~\ref{tab:main_5_10} and Table~\ref{tab:agents_num}, CoELA occasionally achieves a lower caught rate than CoSaR. However, this does not reflect safer navigation. Instead, CoELA tends to spend much more time in initial locations, where agents are not exposed to sentinels. While this reduces the caught rate, it also delays coordination and movement, leading to a lower overall success rate.

\subsection{Ablation Study}

To systematically evaluate the effectiveness of different components in our method, we conduct an ablation study by removing each of the following four components:
\begin{enumerate}
    \item Route Refinement: The retry limit for route refinement attempts is set to zero.
    \item Message Analyzer: The message analyzer is not invoked when a new message is received. Consequently, the ETA map and the list of known poses are not stored, and the summarized opinions of each agent are unavailable.
    \item Spatial Memory: Agents are disabled from accessing spatial memory when making decisions. As a result, they cannot utilize the ETA map or the list of known poses.
    \item Emergency Avoidance: Agents do not perform emergency avoidance when being too close to any sentinel. This leads to bolder navigation behavior.
\end{enumerate}
The results are shown in Table~\ref{tab:ablate}. The success rate consistently decreases as these components are removed, with the largest drop occurring when the ETA map and the list of known poses are unavailable due to disabling the Analyzer and Spatial Memory. This mainly reduces the agents’ ability to identify suitable meeting locations and increases their exposure to sentinel threats.

\begin{table}[t]
\centering
\small
\caption{\textbf{Ablation Results.} We report the results under the setting of 5 agents and 10 stationary sentinels over 14 scenes and 6 runs here.}
\resizebox{0.9\linewidth}{!}{
\begin{tabular}{lcccccc}
\toprule
Method  &   \makecell{Success\\Rate\,$\uparrow$} & \makecell{Caught\\Rate$\downarrow$} 
& \makecell{Detection\\Rate$\downarrow$} & \makecell{Time.$\downarrow$} & \makecell{Dist.$\downarrow$}\\
\midrule
CoSaR (Ours) &  \textbf{32.14} & \textbf{21.43} & 1.27 & \textbf{1194.45} & 1552.45\\
without Route Refinement &  26.19 & 28.80 & 1.18 & 1243.31 & \textbf{1446.25}\\
without Message Analyzer &  19.26 & 31.38 & 1.29 & 1310.20 & 1488.76\\
without Spatial Memory &  21.42 & 31.19 & 1.27 & 1283.70 & 1489.28\\
without Emergency Avoidance &  24.05 & 31.95 & \textbf{1.15} & 1268.58 & 1513.40\\
\bottomrule
\end{tabular}
}
\label{tab:ablate}
\end{table}

\subsection{Backbone Comparison}

To compare the performance of our method across different Foundation Model backbones, we additionally experimented using \texttt{Qwen3-VL-30B-A3B-Instruct} with our method. As shown in Table~\ref{tab:qwen}, our method with \texttt{Qwen} backbone also achieves competing performance, demonstrating the robustness and effectiveness of our methods across different backbones.

\begin{table}[t]
\centering
\small
\caption{\textbf{Performance Comparison Across Foundation Model Backbones.} We report the results under the setting of 5 agents and 5 stationary sentinels over 14 scenes and 2 runs here.}
\vspace{-3mm}
\resizebox{0.7\linewidth}{!}{
\begin{tabular}{lcccccc}
\toprule
Method  &   \makecell{Success\\Rate\,$\uparrow$} & \makecell{Caught\\Rate$\downarrow$} 
& \makecell{Detection\\Rate$\downarrow$} & \makecell{Time.$\downarrow$} & \makecell{Dist.$\downarrow$}\\
\midrule
\texttt{gpt-4o} &  67.85 & 5.00 & 0.57 & 950.5 & 1172.15\\
\texttt{Qwen3-VL-30B} &  60.71 & 7.69 & 0.77 & 1013.38 & 1292.97\\
\bottomrule
\end{tabular}
}
\label{tab:qwen}
\end{table}

\section{Discussion}

Beyond robotics applications—such as search-and-rescue, human-aware navigation, and autonomous delivery under dynamic constraints—this benchmark is also motivated by large-scale multi-agent environments, including game-like settings where agents must coordinate while avoiding hostile or visibility-constrained entities. In this formulation, sentinels abstract dynamic hazards such as guards, adversaries, surveillance regions, or unsafe zones, requiring agents to jointly reason about risk, visibility, and coordination.

More broadly, the benchmark is designed to study cooperative spatial reasoning and communication under partial observability and evolving constraints. It provides a controlled setting for analyzing how agents share information, adapt plans, and maintain coordination in the presence of dynamic, spatially grounded risks.

We acknowledge that such capabilities carry potential dual-use implications. In particular, techniques for coordinated navigation and inference under partial observability could be misused in surveillance-oriented contexts, such as tracking or path prediction. To mitigate these risks, we position the benchmark strictly as a research platform for studying safe and cooperative embodied intelligence. We encourage future work to incorporate safeguards, transparency mechanisms, and domain restrictions when considering real-world deployment.
\section{Conclusion}

In this work, we introduced the Sentinel Challenge as a benchmark for studying embodied multi-agent cooperation under dynamic and adversarial spatial constraints. To address this setting, we proposed CoSaR, a cooperative spatial reasoning and planning framework that equips agents with a dynamic spatial memory for integrating perception, communication, and map-based signals. This unified representation enables agents to share information effectively, reason about evolving spatial structure, and adapt their plans through coordinated replanning. Across 14 diverse city-scale environments involving 3–5 agents, CoSaR consistently outperforms strong baselines, demonstrating the importance of spatially grounded memory and communication in challenging multi-agent navigation tasks. We believe Sentinel Challenge and CoSaR provide a promising foundation for future research on large-scale embodied cooperation, resilient multi-agent planning, and spatially grounded communication.

\section*{Acknowledgement}

This work was supported by NSF IIS-2441250, NSF IIS-2404386 and MURI.


%
%
\bibliographystyle{splncs04}
\bibliography{main}

@String(AAAI = {AAAI})

@article{kahn2021badgr,
  title={Badgr: An autonomous self-supervised learning-based navigation system},
  author={Kahn, Gregory and Abbeel, Pieter and Levine, Sergey},
  journal={IEEE Robotics and Automation Letters},
  volume={6},
  number={2},
  pages={1312--1319},
  year={2021},
  publisher={IEEE}
}

@article{wijmans2019dd,
  title={Dd-ppo: Learning near-perfect pointgoal navigators from 2.5 billion frames},
  author={Wijmans, Erik and Kadian, Abhishek and Morcos, Ari and Lee, Stefan and Essa, Irfan and Parikh, Devi and Savva, Manolis and Batra, Dhruv},
  journal={arXiv preprint arXiv:1911.00357},
  year={2019}
}

@inproceedings{cai2025spatialbot,
  title={Spatialbot: Precise spatial understanding with vision language models},
  author={Cai, Wenxiao and Ponomarenko, Iaroslav and Yuan, Jianhao and Li, Xiaoqi and Yang, Wankou and Dong, Hao and Zhao, Bo},
  booktitle={2025 IEEE International Conference on Robotics and Automation (ICRA)},
  pages={9490--9498},
  year={2025},
  organization={IEEE}
}

@article{fu2024scene,
  title={Scene-llm: Extending language model for 3d visual understanding and reasoning},
  author={Fu, Rao and Liu, Jingyu and Chen, Xilun and Nie, Yixin and Xiong, Wenhan},
  journal={arXiv preprint arXiv:2403.11401},
  year={2024}
}

@article{hong20233d,
  title={3d-llm: Injecting the 3d world into large language models},
  author={Hong, Yining and Zhen, Haoyu and Chen, Peihao and Zheng, Shuhong and Du, Yilun and Chen, Zhenfang and Gan, Chuang},
  journal={Advances in Neural Information Processing Systems},
  volume={36},
  pages={20482--20494},
  year={2023}
}

@inproceedings{wang2024embodiedscan,
  title={Embodiedscan: A holistic multi-modal 3d perception suite towards embodied ai},
  author={Wang, Tai and Mao, Xiaohan and Zhu, Chenming and Xu, Runsen and Lyu, Ruiyuan and Li, Peisen and Chen, Xiao and Zhang, Wenwei and Chen, Kai and Xue, Tianfan and others},
  booktitle={Proceedings of the IEEE/CVF Conference on Computer Vision and Pattern Recognition},
  pages={19757--19767},
  year={2024}
}

@article{dijkstra1959note,
  title={A note on two problems in connexion with graphs},
  author={Dijkstra, Edsger W.},
  journal={Numerische Mathematik},
  volume={1},
  number={1},
  pages={269--271},
  year={1959},
  publisher={Springer}
}

@inproceedings{radford2021learning,
  title={Learning transferable visual models from natural language supervision},
  author={Radford, Alec and Kim, Jong Wook and Hallacy, Chris and Ramesh, Aditya and Goh, Gabriel and Agarwal, Sandhini and Sastry, Girish and Askell, Amanda and Mishkin, Pamela and Clark, Jack and others},
  booktitle={International conference on machine learning},
  pages={8748--8763},
  year={2021},
  organization={PmLR}
}

@inproceedings{yang20253d,
  title={3D-mem: 3D scene memory for embodied exploration and reasoning},
  author={Yang, Yuncong and Yang, Han and Zhou, Jiachen and Chen, Peihao and Zhang, Hongxin and Du, Yilun and Gan, Chuang},
  booktitle={Proceedings of the Computer Vision and Pattern Recognition Conference},
  pages={17294--17303},
  year={2025}
}

@inproceedings{khandelwal2022simple,
  title={Simple but effective: Clip embeddings for embodied ai},
  author={Khandelwal, Apoorv and Weihs, Luca and Mottaghi, Roozbeh and Kembhavi, Aniruddha},
  booktitle={Proceedings of the IEEE/CVF Conference on Computer Vision and Pattern Recognition},
  pages={14829--14838},
  year={2022}
}

@inproceedings{sridhar2024nomad,
  title={Nomad: Goal masked diffusion policies for navigation and exploration},
  author={Sridhar, Ajay and Shah, Dhruv and Glossop, Catherine and Levine, Sergey},
  booktitle={2024 IEEE International Conference on Robotics and Automation (ICRA)},
  pages={63--70},
  year={2024},
  organization={IEEE}
}

@article{shah2023vint,
  title={ViNT: A foundation model for visual navigation},
  author={Shah, Dhruv and Sridhar, Ajay and Dashora, Nitish and Stachowicz, Kyle and Black, Kevin and Hirose, Noriaki and Levine, Sergey},
  journal={arXiv preprint arXiv:2306.14846},
  year={2023}
}

@article{morales2009autonomous,
  title={Autonomous robot navigation in outdoor cluttered pedestrian walkways},
  author={Morales, Yoichi and Carballo, Alexander and Takeuchi, Eijiro and Aburadani, Atsushi and Tsubouchi, Takashi},
  journal={Journal of Field Robotics},
  volume={26},
  number={8},
  pages={609--635},
  year={2009},
  publisher={Wiley Online Library}
}

@inproceedings{kummerle2013navigation,
  title={A navigation system for robots operating in crowded urban environments},
  author={K{\"u}mmerle, Rainer and Ruhnke, Michael and Steder, Bastian and Stachniss, Cyrill and Burgard, Wolfram},
  booktitle={2013 IEEE International Conference on Robotics and Automation},
  pages={3225--3232},
  year={2013},
  organization={IEEE}
}

@inproceedings{liu2025citywalker,
  title={Citywalker: Learning embodied urban navigation from web-scale videos},
  author={Liu, Xinhao and Li, Jintong and Jiang, Yicheng and Sujay, Niranjan and Yang, Zhicheng and Zhang, Juexiao and Abanes, John and Zhang, Jing and Feng, Chen},
  booktitle={Proceedings of the Computer Vision and Pattern Recognition Conference},
  pages={6875--6885},
  year={2025}
}

@article{anderson2018evaluation,
  title={On evaluation of embodied navigation agents},
  author={Anderson, Peter and Chang, Angel and Chaplot, Devendra Singh and Dosovitskiy, Alexey and Gupta, Saurabh and Koltun, Vladlen and Kosecka, Jana and Malik, Jitendra and Mottaghi, Roozbeh and Savva, Manolis and others},
  journal={arXiv preprint arXiv:1807.06757},
  year={2018}
}

@inproceedings{yang2025thinking,
  title={Thinking in space: How multimodal large language models see, remember, and recall spaces},
  author={Yang, Jihan and Yang, Shusheng and Gupta, Anjali W and Han, Rilyn and Fei-Fei, Li and Xie, Saining},
  booktitle={Proceedings of the Computer Vision and Pattern Recognition Conference},
  pages={10632--10643},
  year={2025}
}

@inproceedings{zheng2024towards,
  title={Towards learning a generalist model for embodied navigation},
  author={Zheng, Duo and Huang, Shijia and Zhao, Lin and Zhong, Yiwu and Wang, Liwei},
  booktitle={Proceedings of the IEEE/CVF Conference on Computer Vision and Pattern Recognition},
  pages={13624--13634},
  year={2024}
}

@inproceedings{song2023llm,
  title={Llm-planner: Few-shot grounded planning for embodied agents with large language models},
  author={Song, Chan Hee and Wu, Jiaman and Washington, Clayton and Sadler, Brian M and Chao, Wei-Lun and Su, Yu},
  booktitle={Proceedings of the IEEE/CVF international conference on computer vision},
  pages={2998--3009},
  year={2023}
}

@article{huang2022visual,
  title={Visual language maps for robot navigation},
  author={Huang, Chenguang and Mees, Oier and Zeng, Andy and Burgard, Wolfram},
  journal={arXiv preprint arXiv:2210.05714},
  year={2022}
}

@inproceedings{shah2023lm,
  title={Lm-nav: Robotic navigation with large pre-trained models of language, vision, and action},
  author={Shah, Dhruv and Osi{\'n}ski, B{\l}a{\.z}ej and Levine, Sergey and others},
  booktitle={Conference on robot learning},
  pages={492--504},
  year={2023},
  organization={PMLR}
}

@inproceedings{shah2021ving,
  title={Ving: Learning open-world navigation with visual goals},
  author={Shah, Dhruv and Eysenbach, Benjamin and Kahn, Gregory and Rhinehart, Nicholas and Levine, Sergey},
  booktitle={2021 IEEE International Conference on Robotics and Automation (ICRA)},
  pages={13215--13222},
  year={2021},
  organization={IEEE}
}

@article{majumdar2022zson,
  title={Zson: Zero-shot object-goal navigation using multimodal goal embeddings},
  author={Majumdar, Arjun and Aggarwal, Gunjan and Devnani, Bhavika and Hoffman, Judy and Batra, Dhruv},
  journal={Advances in Neural Information Processing Systems},
  volume={35},
  pages={32340--32352},
  year={2022}
}

@article{gadre2022clip,
  title={Clip on wheels: Zero-shot object navigation as object localization and exploration},
  author={Gadre, Samir Yitzhak and Wortsman, Mitchell and Ilharco, Gabriel and Schmidt, Ludwig and Song, Shuran},
  journal={arXiv preprint arXiv:2203.10421},
  volume={3},
  number={4},
  pages={7},
  year={2022}
}

@article{kim2024openvla,
  title={Openvla: An open-source vision-language-action model},
  author={Kim, Moo Jin and Pertsch, Karl and Karamcheti, Siddharth and Xiao, Ted and Balakrishna, Ashwin and Nair, Suraj and Rafailov, Rafael and Foster, Ethan and Lam, Grace and Sanketi, Pannag and others},
  journal={arXiv preprint arXiv:2406.09246},
  year={2024}
}

@article{mon2025embodied,
  title={Embodied large language models enable robots to complete complex tasks in unpredictable environments},
  author={Mon-Williams, Ruaridh and Li, Gen and Long, Ran and Du, Wenqian and Lucas, Christopher G},
  journal={Nature Machine Intelligence},
  pages={1--10},
  year={2025},
  publisher={Nature Publishing Group UK London}
}

@article{zhang2024combo,
  title={COMBO: compositional world models for embodied multi-agent cooperation},
  author={Zhang, Hongxin and Wang, Zeyuan and Lyu, Qiushi and Zhang, Zheyuan and Chen, Sunli and Shu, Tianmin and Dariush, Behzad and Lee, Kwonjoon and Du, Yilun and Gan, Chuang},
  journal={arXiv preprint arXiv:2404.10775},
  year={2024}
}

@inproceedings{zitkovich2023rt,
  title={Rt-2: Vision-language-action models transfer web knowledge to robotic control},
  author={Zitkovich, Brianna and Yu, Tianhe and Xu, Sichun and Xu, Peng and Xiao, Ted and Xia, Fei and Wu, Jialin and Wohlhart, Paul and Welker, Stefan and Wahid, Ayzaan and others},
  booktitle={Conference on Robot Learning},
  pages={2165--2183},
  year={2023},
  organization={PMLR}
}

@inproceedings{jiang2023vima,
  title     = {VIMA: General Robot Manipulation with Multimodal Prompts},
  author    = {Yunfan Jiang and Agrim Gupta and Zichen Zhang and Guanzhi Wang and Yongqiang Dou and Yanjun Chen and Li Fei-Fei and Anima Anandkumar and Yuke Zhu and Linxi Fan},
  booktitle = {Fortieth International Conference on Machine Learning},
  year      = {2023}
}

@inproceedings{zu2025collaborative,
  title={Collaborative Tree Search for Enhancing Embodied Multi-Agent Collaboration},
  author={Zu, Lizheng and Lin, Lin and Fu, Song and Zhao, Na and Zhou, Pan},
  booktitle={Proceedings of the Computer Vision and Pattern Recognition Conference},
  pages={29513--29522},
  year={2025}
}

@misc{openai2023gpt4,
      title={GPT-4 Technical Report}, 
      author={OpenAI},
      year={2023},
      eprint={2303.08774},
      archivePrefix={arXiv},
      primaryClass={cs.CL}
}

@misc{zhang2023building,
      title={Building Cooperative Embodied Agents Modularly with Large Language Models}, 
      author={Hongxin Zhang and Weihua Du and Jiaming Shan and Qinhong Zhou and Yilun Du and Joshua B. Tenenbaum and Tianmin Shu and Chuang Gan},
      year={2023},
      eprint={2307.02485},
      archivePrefix={arXiv},
      primaryClass={cs.AI}
}

@article{xi2023rise,
  title={The Rise and Potential of Large Language Model Based Agents: A Survey},
  author={Xi, Zhiheng and Chen, Wenxiang and Guo, Xin and He, Wei and Ding, Yiwen and Hong, Boyang and Zhang, Ming and Wang, Junzhe and Jin, Senjie and Zhou, Enyu and others},
  journal={arXiv preprint arXiv:2309.07864},
  year={2023}
}

@article{wang2023survey,
  title={A survey on large language model based autonomous agents},
  author={Wang, Lei and Ma, Chen and Feng, Xueyang and Zhang, Zeyu and Yang, Hao and Zhang, Jingsen and Chen, Zhiyuan and Tang, Jiakai and Chen, Xu and Lin, Yankai and others},
  journal={arXiv preprint arXiv:2308.11432},
  year={2023}
}

@article{sumers2023cognitive,
  title={Cognitive architectures for language agents},
  author={Sumers, Theodore and Yao, Shunyu and Narasimhan, Karthik and Griffiths, Thomas L},
  journal={arXiv preprint arXiv:2309.02427},
  year={2023}
}

@article{wang2023voyager,
  title={Voyager: An open-ended embodied agent with large language models},
  author={Wang, Guanzhi and Xie, Yuqi and Jiang, Yunfan and Mandlekar, Ajay and Xiao, Chaowei and Zhu, Yuke and Fan, Linxi and Anandkumar, Anima},
  journal={arXiv preprint arXiv:2305.16291},
  year={2023}
}

@article{sharma2021skill,
  title={Skill induction and planning with latent language},
  author={Sharma, Pratyusha and Torralba, Antonio and Andreas, Jacob},
  journal={arXiv preprint arXiv:2110.01517},
  year={2021}
}

@article{ahn2022can,
  title={Do as i can, not as i say: Grounding language in robotic affordances},
  author={Ahn, Michael and Brohan, Anthony and Brown, Noah and Chebotar, Yevgen and Cortes, Omar and David, Byron and Finn, Chelsea and Gopalakrishnan, Keerthana and Hausman, Karol and Herzog, Alex and others},
  journal={arXiv preprint arXiv:2204.01691},
  year={2022}
}

@inproceedings{wang2023describe,
  title={Describe, explain, plan and select: interactive planning with LLMs enables open-world multi-task agents},
  author={Wang, Zihao and Cai, Shaofei and Chen, Guanzhou and Liu, Anji and Ma, Xiaojian and Liang, Yitao},
  booktitle={Thirty-seventh Conference on Neural Information Processing Systems},
  year={2023}
}

@inproceedings{
wang2023jarvis,
title={{JARVIS}-1: Open-world Multi-task Agents with Memory-Augmented Multimodal Language Models},
author={Zihao Wang and Shaofei Cai and Anji Liu and Xiaojian Ma and Yitao Liang},
booktitle={Second Agent Learning in Open-Endedness Workshop},
year={2023},
url={https://openreview.net/forum?id=xzPkZyHlOW}
}

@inproceedings{szot2023adaptive,
  title={Adaptive coordination in social embodied rearrangement},
  author={Szot, Andrew and Jain, Unnat and Batra, Dhruv and Kira, Zsolt and Desai, Ruta and Rai, Akshara},
  booktitle={International Conference on Machine Learning},
  pages={33365--33380},
  year={2023},
  organization={PMLR}
}

@inproceedings{jain2020cordial,
  title={A cordial sync: Going beyond marginal policies for multi-agent embodied tasks},
  author={Jain, Unnat and Weihs, Luca and Kolve, Eric and Farhadi, Ali and Lazebnik, Svetlana and Kembhavi, Aniruddha and Schwing, Alexander},
  booktitle={Computer Vision--ECCV 2020: 16th European Conference, Glasgow, UK, August 23--28, 2020, Proceedings, Part V 16},
  pages={471--490},
  year={2020},
  organization={Springer}
}

@inproceedings{puigwatch,
  title={Watch-And-Help: A Challenge for Social Perception and Human-AI Collaboration},
  author={Puig, Xavier and Shu, Tianmin and Li, Shuang and Wang, Zilin and Liao, Yuan-Hong and Tenenbaum, Joshua B and Fidler, Sanja and Torralba, Antonio},
  year={2021},
  booktitle={International Conference on Learning Representations}
}

@article{puig2023habitat,
  title={Habitat 3.0: A co-habitat for humans, avatars and robots},
  author={Puig, Xavier and Undersander, Eric and Szot, Andrew and Cote, Mikael Dallaire and Yang, Tsung-Yen and Partsey, Ruslan and Desai, Ruta and Clegg, Alexander William and Hlavac, Michal and Min, So Yeon and others},
  journal={arXiv preprint arXiv:2310.13724},
  year={2023}
}

@article{park2023generative,
  title={Generative agents: Interactive simulacra of human behavior},
  author={Park, Joon Sung and O'Brien, Joseph C and Cai, Carrie J and Morris, Meredith Ringel and Liang, Percy and Bernstein, Michael S},
  journal={arXiv preprint arXiv:2304.03442},
  year={2023}
}

@article{mandi2023roco,
  title={Roco: Dialectic multi-robot collaboration with large language models},
  author={Mandi, Zhao and Jain, Shreeya and Song, Shuran},
  journal={arXiv preprint arXiv:2307.04738},
  year={2023}
}

@article{stone2000multiagent,
  title={Multiagent systems: A survey from a machine learning perspective},
  author={Stone, Peter and Veloso, Manuela},
  journal={Autonomous Robots},
  volume={8},
  pages={345--383},
  year={2000},
  publisher={Springer}
}

@article{bernstein2002complexity,
  title={The complexity of decentralized control of Markov decision processes},
  author={Bernstein, Daniel S and Givan, Robert and Immerman, Neil and Zilberstein, Shlomo},
  journal={Mathematics of operations research},
  volume={27},
  number={4},
  pages={819--840},
  year={2002},
  publisher={INFORMS}
}

@inproceedings{spaan2006decentralized,
  title={Decentralized planning under uncertainty for teams of communicating agents},
  author={Spaan, Matthijs TJ and Gordon, Geoffrey J and Vlassis, Nikos},
  booktitle={Proceedings of the fifth international joint conference on Autonomous agents and multiagent systems},
  pages={249--256},
  year={2006}
}

@inproceedings{
Baker2020Emergent,
title={Emergent Tool Use From Multi-Agent Autocurricula},
author={Bowen Baker and Ingmar Kanitscheider and Todor Markov and Yi Wu and Glenn Powell and Bob McGrew and Igor Mordatch},
booktitle={International Conference on Learning Representations},
year={2020},
url={https://openreview.net/forum?id=SkxpxJBKwS}
}

@article{lowe2017multi,
  title={Multi-agent actor-critic for mixed cooperative-competitive environments},
  author={Lowe, Ryan and Tamar, Aviv and Harb, Jean and Pieter Abbeel, OpenAI and Mordatch, Igor},
  journal={Advances in neural information processing systems},
  volume={30},
  year={2017}
}

@inproceedings{samvelyan2019starcraft,
  title={The StarCraft Multi-Agent Challenge},
  author={Samvelyan, Mikayel and Rashid, Tabish and Schroeder de Witt, Christian and Farquhar, Gregory and Nardelli, Nantas and Rudner, Tim GJ and Hung, Chia-Man and Torr, Philip HS and Foerster, Jakob and Whiteson, Shimon},
  booktitle={Proceedings of the 18th International Conference on Autonomous Agents and MultiAgent Systems},
  pages={2186--2188},
  year={2019}
}

@article{carroll2019utility,
  title={On the utility of learning about humans for human-ai coordination},
  author={Carroll, Micah and Shah, Rohin and Ho, Mark K and Griffiths, Tom and Seshia, Sanjit and Abbeel, Pieter and Dragan, Anca},
  journal={Advances in neural information processing systems},
  volume={32},
  year={2019}
}

@article{jaderberg2019human,
  title={Human-level performance in 3D multiplayer games with population-based reinforcement learning},
  author={Jaderberg, Max and Czarnecki, Wojciech M and Dunning, Iain and Marris, Luke and Lever, Guy and Castaneda, Antonio Garcia and Beattie, Charles and Rabinowitz, Neil C and Morcos, Ari S and Ruderman, Avraham and others},
  journal={Science},
  volume={364},
  number={6443},
  pages={859--865},
  year={2019},
  publisher={American Association for the Advancement of Science}
}

@article{suarez2019neural,
  title={Neural MMO: A massively multiagent game environment for training and evaluating intelligent agents},
  author={Suarez, Joseph and Du, Yilun and Isola, Phillip and Mordatch, Igor},
  journal={arXiv preprint arXiv:1903.00784},
  year={2019}
}

@article{bard2020hanabi,
  title={The hanabi challenge: A new frontier for ai research},
  author={Bard, Nolan and Foerster, Jakob N and Chandar, Sarath and Burch, Neil and Lanctot, Marc and Song, H Francis and Parisotto, Emilio and Dumoulin, Vincent and Moitra, Subhodeep and Hughes, Edward and others},
  journal={Artificial Intelligence},
  volume={280},
  pages={103216},
  year={2020},
  publisher={Elsevier}
}

@article{wen2022multi,
  title={Multi-agent reinforcement learning is a sequence modeling problem},
  author={Wen, Muning and Kuba, Jakub and Lin, Runji and Zhang, Weinan and Wen, Ying and Wang, Jun and Yang, Yaodong},
  journal={Advances in Neural Information Processing Systems},
  volume={35},
  pages={16509--16521},
  year={2022}
}

@article{amato2019modeling,
  title={Modeling and planning with macro-actions in decentralized POMDPs},
  author={Amato, Christopher and Konidaris, George and Kaelbling, Leslie P and How, Jonathan P},
  journal={Journal of Artificial Intelligence Research},
  volume={64},
  pages={817--859},
  year={2019}
}

@misc{zhou2024hazard,
      title={HAZARD Challenge: Embodied Decision Making in Dynamically Changing Environments}, 
      author={Qinhong Zhou and Sunli Chen and Yisong Wang and Haozhe Xu and Weihua Du and Hongxin Zhang and Yilun Du and Joshua B. Tenenbaum and Chuang Gan},
      year={2024},
      eprint={2401.12975},
      archivePrefix={arXiv},
      primaryClass={cs.CV}
}

@article{kolve2017ai2,
  title={Ai2-thor: An interactive 3d environment for visual ai},
  author={Kolve, Eric and Mottaghi, Roozbeh and Han, Winson and VanderBilt, Eli and Weihs, Luca and Herrasti, Alvaro and Deitke, Matt and Ehsani, Kiana and Gordon, Daniel and Zhu, Yuke and others},
  journal={arXiv preprint arXiv:1712.05474},
  year={2017}
}

@inproceedings{savva2019habitat,
  title={Habitat: A platform for embodied ai research},
  author={Savva, Manolis and Kadian, Abhishek and Maksymets, Oleksandr and Zhao, Yili and Wijmans, Erik and Jain, Bhavana and Straub, Julian and Liu, Jia and Koltun, Vladlen and Malik, Jitendra and others},
  booktitle={Proceedings of the IEEE/CVF international conference on computer vision},
  pages={9339--9347},
  year={2019}
}

@inproceedings{xia2018gibson,
  title={Gibson env: Real-world perception for embodied agents},
  author={Xia, Fei and Zamir, Amir R and He, Zhiyang and Sax, Alexander and Malik, Jitendra and Savarese, Silvio},
  booktitle={Proceedings of the IEEE conference on computer vision and pattern recognition},
  pages={9068--9079},
  year={2018}
}

@article{misra2018mapping,
  title={Mapping instructions to actions in 3d environments with visual goal prediction},
  author={Misra, Dipendra and Bennett, Andrew and Blukis, Valts and Niklasson, Eyvind and Shatkhin, Max and Artzi, Yoav},
  journal={arXiv preprint arXiv:1809.00786},
  year={2018}
}

@inproceedings{li2023behavior,
  title={Behavior-1k: A benchmark for embodied ai with 1,000 everyday activities and realistic simulation},
  author={Li, Chengshu and Zhang, Ruohan and Wong, Josiah and Gokmen, Cem and Srivastava, Sanjana and Mart{\'\i}n-Mart{\'\i}n, Roberto and Wang, Chen and Levine, Gabrael and Lingelbach, Michael and Sun, Jiankai and others},
  booktitle={Conference on Robot Learning},
  pages={80--93},
  year={2023},
  organization={PMLR}
}

@inproceedings{padmakumar2022teach,
  title={Teach: Task-driven embodied agents that chat},
  author={Padmakumar, Aishwarya and Thomason, Jesse and Shrivastava, Ayush and Lange, Patrick and Narayan-Chen, Anjali and Gella, Spandana and Piramuthu, Robinson and Tur, Gokhan and Hakkani-Tur, Dilek},
  booktitle={Proceedings of the AAAI Conference on Artificial Intelligence},
  volume={36},
  pages={2017--2025},
  year={2022}
}

@inproceedings{xiang2020sapien,
  title={Sapien: A simulated part-based interactive environment},
  author={Xiang, Fanbo and Qin, Yuzhe and Mo, Kaichun and Xia, Yikuan and Zhu, Hao and Liu, Fangchen and Liu, Minghua and Jiang, Hanxiao and Yuan, Yifu and Wang, He and others},
  booktitle={Proceedings of the IEEE/CVF Conference on Computer Vision and Pattern Recognition},
  pages={11097--11107},
  year={2020}
}

@article{liu2023grounding,
  title={Grounding dino: Marrying dino with grounded pre-training for open-set object detection},
  author={Liu, Shilong and Zeng, Zhaoyang and Ren, Tianhe and Li, Feng and Zhang, Hao and Yang, Jie and Jiang, Qing and Li, Chunyuan and Yang, Jianwei and Su, Hang and others},
  journal={arXiv preprint arXiv:2303.05499},
  year={2023}
}

@misc{ravi2024sam2segmentimages,
      title={SAM 2: Segment Anything in Images and Videos}, 
      author={Nikhila Ravi and Valentin Gabeur and Yuan-Ting Hu and Ronghang Hu and Chaitanya Ryali and Tengyu Ma and Haitham Khedr and Roman Rädle and Chloe Rolland and Laura Gustafson and Eric Mintun and Junting Pan and Kalyan Vasudev Alwala and Nicolas Carion and Chao-Yuan Wu and Ross Girshick and Piotr Dollár and Christoph Feichtenhofer},
      year={2024},
      eprint={2408.00714},
      archivePrefix={arXiv},
      primaryClass={cs.CV},
      url={https://arxiv.org/abs/2408.00714}, 
}

@article{guo2025deepseek,
  title={DeepSeek-R1: Incentivizing Reasoning Capability in LLMs via Reinforcement Learning},
  author={Guo, Daya and Yang, Dejian and Zhang, Haowei and Song, Junxiao and Zhang, Ruoyu and Xu, Runxin and Zhu, Qihao and Ma, Shirong and Wang, Peiyi and Bi, Xiao and others},
  journal={arXiv preprint arXiv:2501.12948},
  year={2025}
}

@article{liu2024capo,
  title={CaPo: Cooperative Plan Optimization for Efficient Embodied Multi-Agent Cooperation},
  author={Liu, Jie and Zhou, Pan and Du, Yingjun and Tan, Ah-Hwee and Snoek, Cees GM and Sonke, Jan-Jakob and Gavves, Efstratios},
  journal={arXiv preprint arXiv:2411.04679},
  year={2024}
}

@article{zhou2025virtual,
  title={Virtual community: An open world for humans, robots, and society},
  author={Zhou, Qinhong and Zhang, Hongxin and Lin, Xiangye and Zhang, Zheyuan and Chen, Yutian and Liu, Wenjun and Zhang, Zunzhe and Chen, Sunli and Fang, Lixing and Lyu, Qiushi and others},
  journal={arXiv preprint arXiv:2508.14893},
  year={2025}
}

@article{zhang2025ella,
  title={Ella: Embodied social agents with lifelong memory},
  author={Zhang, Hongxin and Zhang, Zheyuan and Wang, Zeyuan and Zhang, Zunzhe and Fang, Lixing and Zhou, Qinhong and Gan, Chuang},
  journal={arXiv preprint arXiv:2506.24019},
  year={2025}
}

\appendix

\tcbset{
    mypanel/.style={
        enhanced,
        colback=white,
        colframe=blue!60!black,
        colbacktitle=blue!70!black,
        coltitle=white,
        fonttitle=\bfseries,
        title=#1,
        boxrule=1pt,
        arc=3mm,
        attach boxed title to top left={xshift=4mm,yshift*=-2mm},
        boxed title style={
            frame code={},
            interior engine=empty,
            top=2pt, bottom=2pt,
        }
    }
}

\lstset{
basicstyle=\ttfamily\small,
breaklines=true,
frame=single
}

\section{Additional Details on the Sentinel Challenge}

The navigation component of the \textsc{Sentinel} challenge is fundamentally built upon the map tool. Since this tool provides the structural support for waypoint routing, obstacle reasoning, and global path computation, we describe its construction and internal representations in detail.

The map tool consists of several core components that collectively define the navigable space of the environment.

\subsection{Waypoints and Connectivity Graph}

Waypoints serve as discrete, navigable anchors placed throughout the city's road network. Each waypoint corresponds to a location in the simulator, and together they form a graph that supports efficient navigation queries.

\paragraph{Waypoint placement.}  
We begin by retrieving all road segments using a built-in feature of \textsc{Virtual Community}, which exposes the complete road topology of the simulated city. Along every road segment, we sample points at a fixed interval of $7$ meters. 

\begin{figure}[t]
    \centering
    \includegraphics[width=1.0\linewidth]{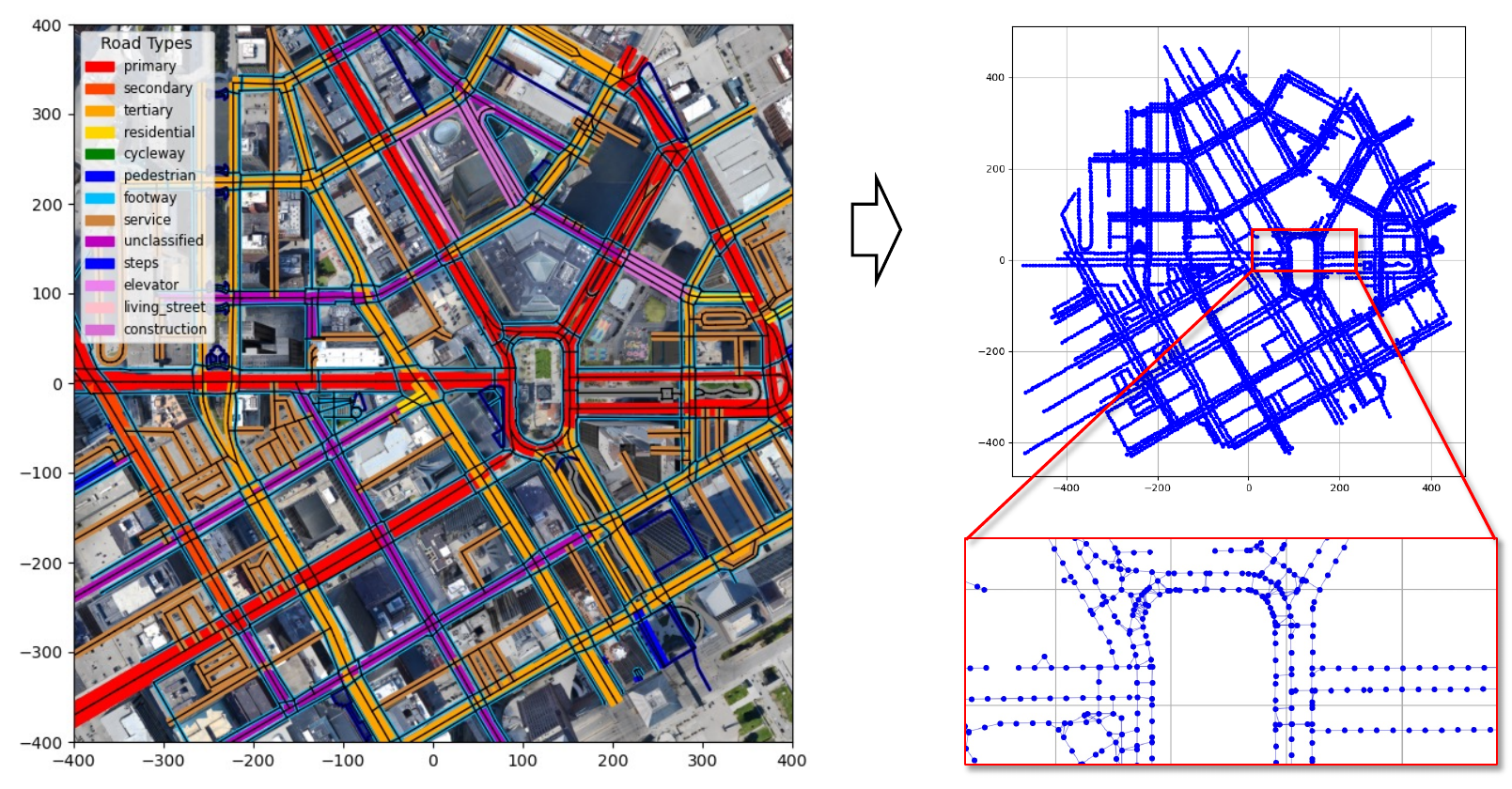}
    \caption{\textbf{Waypoints in the Map Tool}. The left image shows the road annotations provided by the Virtual Community dataset. Based on these annotations, we further generate waypoints and connect them to construct our Map Tool.}
    \label{fig:map_tool}
\end{figure}

\paragraph{Graph connectivity.}  
After all waypoints have been placed, we construct an undirected graph by connecting any two waypoints whose Euclidean distance is at most $7$ meters, as shown in Figure~\ref{fig:map_tool}. This procedure ensures that consecutive samples along the same road are linked, and it naturally captures intersection structures where multiple roads meet within this distance threshold.

The resulting waypoint graph serves as the backbone for high-level map tool functions, enabling efficient computation of shortest paths across the city. 

The \textit{QueryRoute} function of the map tool is further implemented with the Dijkstra algorithm.

\subsection{Coarse Obstacle Grid Map}

In addition to the waypoint graph, the map tool provides a grid-based approximation of all static obstacles (mostly buildings) in the environment.

\paragraph{Obstacle extraction.}  
Obstacle information is obtained from \textsc{Virtual Community}'s built-in scene representation. This data includes buildings, walls, and other non-navigable structures, represented via bounding rectangles. The city map is discretized into a uniform grid, and each grid cell is labeled as either \emph{occupied} or \emph{free}. A grid cell is marked occupied if it intersects with any of the obstacle bounding shapes provided by the environment. Because the underlying obstacle representation often relies on coarse bounding volumes rather than precise geometry, the resulting grid map is an approximation.

It is important to distinguish between the \textbf{coarse obstacle grid map} and the \textbf{occupancy map} constructed by the agents. The coarse obstacle grid map is obtained directly from the built-in database of \textsc{Virtual Community}. As it is only an approximate representation of the environment, its spatial accuracy is insufficient for reliable navigation. Consequently, it is used solely for high-level planning, specifically to construct the \textbf{schematic map} employed during the \textbf{route refinement} process.

In contrast, the occupancy map is built from the agents' visual perception during execution. By integrating depth observations and semantic labels from RGB-D frames, the agent incrementally reconstructs a local representation of the environment that captures obstacles and navigable space at a much finer spatial resolution. This higher level of precision allows the occupancy map to support local path-finding and collision avoidance.

Another key difference lies in how these maps become available. The coarse obstacle grid map provides global structural information from the beginning of the task, offering a rough overview of the environment layout. The occupancy map, however, must be constructed progressively as the agent explores the scene and gathers perceptual observations. As a result, the two maps serve complementary roles: the coarse map provides global but approximate guidance for strategic planning, while the occupancy map supplies accurate local information necessary for safe navigation.

\begin{figure}[t]
    \centering
    \includegraphics[width=0.9\linewidth]{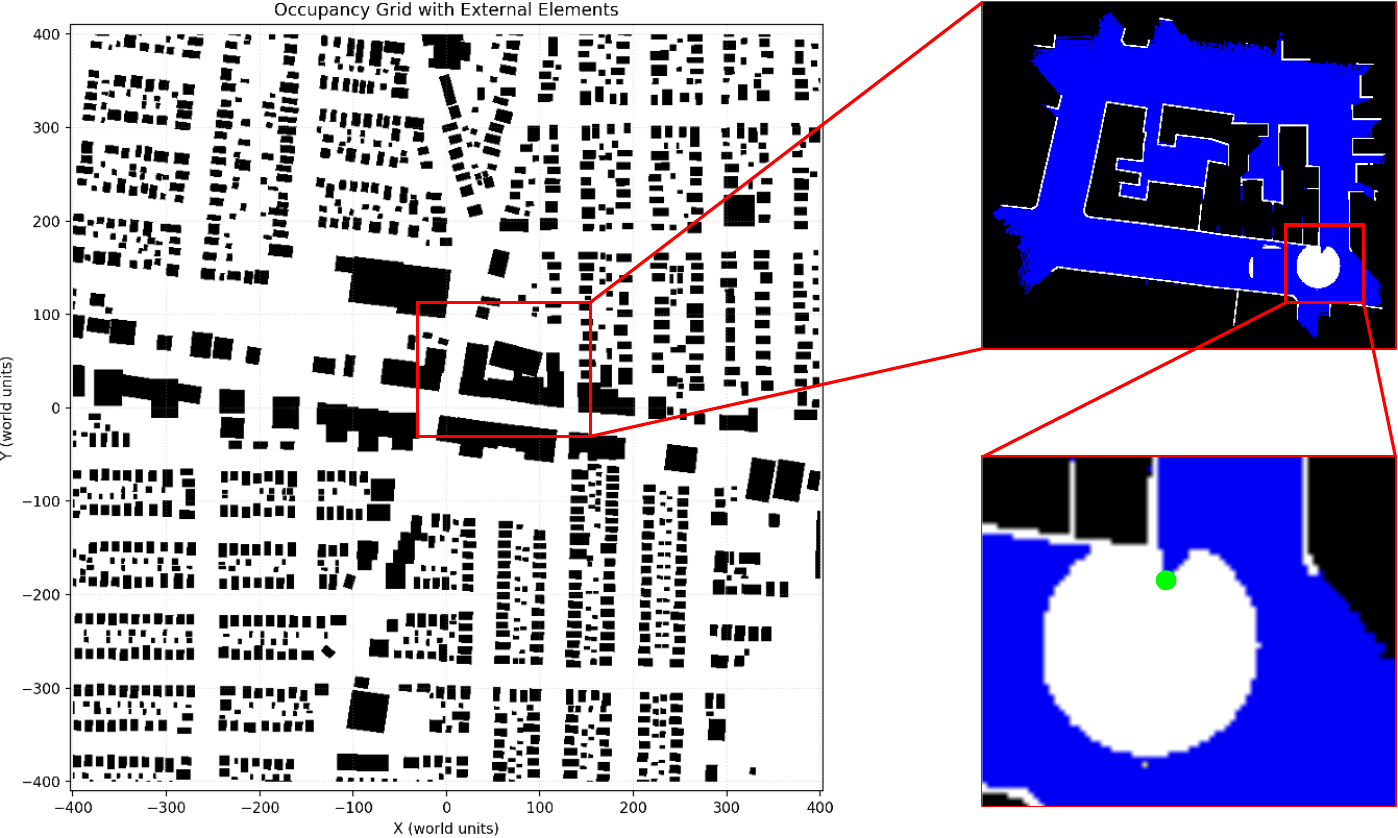}
    \caption{\textbf{Coarse Obstacle Map and Occupancy Map}. Left: the coarse obstacle map of the Long Island scene. Right: the occupancy map observed by an agent, with the agent's position shown as a green dot. The white circle marks a danger zone containing a sentinel. The occupancy map provides significantly finer spatial detail.}
    \label{fig:maps}
\end{figure}

\section{Additional Details on CoSaR}

\subsection{Occupancy Map and Danger Zone}
The occupancy map handles danger zones using the following procedure:

\begin{enumerate}
    \item Each RGB--D frame is accompanied by semantic labels. The agent keeps a set of ``dangerous'' labels corresponding to sentinel-related classes.

    \item For every pixel whose semantic label belongs to this dangerous set, its 3D location is reconstructed from the depth value and camera pose.

    \item Around each reconstructed location, the occupancy map marks nearby cells within a fixed radius as dangerous. Specifically, let $\mathbf{p}_s$ denote the reconstructed sentinel-related point. All cells $\mathbf{p}$ satisfying
    \begin{equation}
        \|\mathbf{p}-\mathbf{p}_s\| \le r
    \end{equation}
    are marked as dangerous, where $r$ is the danger-zone radius. In our experiments, $r$ is set to $10$.

    \item To prevent the agent from being enclosed by newly created danger zones-which may interfere with A* navigation-we restrict danger labeling using a distance-difference constraint relative to the agent position. Let $\mathbf{p}_a$ denote the agent position. A cell $\mathbf{p}$ is marked as dangerous only if
    \begin{equation}
        d(\mathbf{p}, \mathbf{p}_s) - d(\mathbf{p}, \mathbf{p}_a) < \tau ,
    \end{equation}
    where $d(\cdot,\cdot)$ denotes Euclidean distance and $\tau$ is a threshold determined by the agent-sentinel distance. The boundary defined by
    \begin{equation}
        d(\mathbf{p}, \mathbf{p}_s) - d(\mathbf{p}, \mathbf{p}_a) = \tau
    \end{equation}
    forms a hyperbolic curve, ensuring that the generated danger region expands primarily away from the agent rather than enclosing it.

    \item Danger zones are updated continuously. If a region previously marked as dangerous becomes visible again in a later frame and no dangerous labels are detected, the corresponding cells are cleared and restored to a safe state. Consequently, the map reflects currently observed threats and automatically removes outdated danger markings.
\end{enumerate}

\subsection{Emergency avoidance}

To reduce the risk of being caught by nearby sentinels during communication or route refinement, CoSaR agents incorporate an emergency‐avoidance mechanism. The procedure operates as follows:

\begin{enumerate}
    \item Emergency avoidance is triggered whenever:
    \begin{enumerate}
        \item the agent receives a warning signal from any sentinel, or
        \item the agent visually detects a sentinel within a certain threshold.
    \end{enumerate}

    \item Once triggered, the agent performs an \emph{emergency avoidance target selection} and moves toward the target for $10$ consecutive steps, away from the nearest sentinels.

    \item After completing these $10$ steps, the agent scans its surroundings to confirm that no sentinel is still in close proximity. If the area is clear, the agent resumes its original task.
\end{enumerate}

\paragraph{Emergency Avoidance Target Selection.}
When an agent detects nearby sentinels, it performs an emergency avoidance procedure to select a temporary navigation target that moves it away from potential threats. 
First, the agent identifies all sentinels within a predefined distance threshold (40 meters) and treats them as immediate threats. 
The agent then retrieves the local occupancy map from the spatial memory module, which encodes unknown cells, obstacles, and navigable space. 
Only positions that lie within the map bounds and are not labeled as obstacles or invalid regions are considered as candidate coordinates.

To evaluate the safety of a candidate position $(x, y)$, the agent computes an avoidance score relative to all nearby sentinels. 
Let $\mathbf{p}_a = (x_a, y_a)$ denote the current agent position and $\mathbf{p}_s = (x_s, y_s)$ denote the position of a sentinel. 
For a candidate coordinate $\mathbf{p} = (x, y)$, the avoidance score is defined as
\begin{equation}
S(\mathbf{p}) = 
\sum_{\mathbf{p}_s \in \mathcal{S}}
\left(
\frac{x - x_a}{x_s - x_a}
+
\frac{y - y_a}{y_s - y_a}
\right),
\end{equation}
where $\mathcal{S}$ denotes the set of nearby sentinels. 
Intuitively, this score measures whether the candidate location lies in the direction opposite to the sentinel relative to the agent. 
Lower scores correspond to positions that move the agent further away from detected threats.

The agent first checks whether any waypoint from its previously planned route satisfies the safety criterion and yields a sufficiently low score. 
If such a waypoint exists, the agent directly selects it as the avoidance target to preserve navigation continuity. 
Otherwise, the agent searches for candidate coordinates within a local window centered at the agent (a $80 \times 80$ region). 
Valid positions with $S(\mathbf{p}) < 0$ are treated as potential avoidance targets, indicating movement away from the sentinel directions.

To introduce robustness and prevent deterministic oscillations, the final target is selected probabilistically. 
Let $\{S_i\}$ denote the avoidance scores of all candidate positions. 
Sampling weights are computed using an exponential transformation
\begin{equation}
w_i = \exp\!\left(-\frac{S_i}{\sigma + \epsilon}\right),
\end{equation}
where $\sigma$ is the standard deviation of the candidate scores and $\epsilon$ is a small constant for numerical stability. 
These weights are normalized into a probability distribution, and one candidate location is sampled accordingly. 
This mechanism biases the selection toward safer positions while maintaining stochastic exploration when multiple feasible directions exist.

\begin{table*}[t]
\centering
\caption{\textbf{Results for 3 agents and 10 sentinels} We report the average score over 14 scenes and 2 runs here. Best performance is shown in \textbf{bold}. The Oracle Centered baseline is excluded from the oracle-perception scenario, since the method itself does not utilize oracle perception.}
\label{tab:main_3_10}
\resizebox{0.9\linewidth}{!}{
\begin{tabular}{l|ccccc|ccccc}
\toprule
& \multicolumn{5}{c|}{\textbf{10 Stationary Sentinels}} 
& \multicolumn{5}{c}{\textbf{10 Patrolling Sentinels}} \\
\cmidrule(r){2-6} \cmidrule(l){7-11}
Method &
\makecell{Succ.\\Rate$\uparrow$} &
\makecell{Caught\\Rate$\downarrow$} &
\makecell{Detect.\\Rate$\downarrow$} &
Time$\downarrow$ &
Dist.$\downarrow$
&
\makecell{Succ.\\Rate$\uparrow$} &
\makecell{Caught\\Rate$\downarrow$} &
\makecell{Detect.\\Rate$\downarrow$} &
Time$\downarrow$ &
Dist.$\downarrow$
\\
\midrule
\multicolumn{11}{c}{\textit{\textbf{Sentinel Challenge}}} \\
\midrule
Oracle Centered
& 21.43 & 40.48 & 1.54 & 1217.79 & 854.31
& 35.71 & 33.33 & \textbf{1.35} & 1128.14 & 879.97
 \\
Oracle Centered w/ DZ
& 21.43 & 40.48 & 2.45 & 1217.79 & 864.39
& 50.00 & 26.19 & 1.92 & \textbf{983.29} & \textbf{796.86}
 \\
MCTS
& 25.00 & 33.33 & 2.96 & 1171.32 & 933.69
& 39.29 & 27.38 & 2.48 & 1112.71 & 935.02
 \\
RoCo
& 28.57 & 27.38 & 1.50 & 1176.93 & \textbf{750.91}
& 39.29 & 26.19 & 2.27 & 1199.39 & 861.23
 \\
CoELA
& 39.29 & \textbf{26.19} & \textbf{1.22} & 1114.46 & 841.43
& 50.00 & \textbf{19.05} & 1.51 & 1109.93 & 909.86
 \\
CoSaR (Ours)
& \textbf{42.86} & 27.38 & 1.49 & \textbf{1040.75} & 817.22
& \textbf{53.57} & 20.24 & 2.32 & 1020.57 & 851.08
 \\
\midrule
\multicolumn{11}{c}{\textit{\textbf{Sentinel Challenge w/ Oracle Perception}}} \\
\midrule
Oracle Centered w/ DZ
& 50.00 & 21.43 & 2.95 & 908.00 & 725.00
& 57.14 & 16.67 & 1.89 & \textbf{892.14} & 710.66
 \\
MCTS
& \textbf{57.14} & 14.29 & 2.56 & 921.07 & 842.37
& 42.86 & 21.43 & 1.61 & 1071.79 & 994.94
 \\
RoCo
& \textbf{57.14} & \textbf{9.52} & \textbf{1.50} & 961.68 & \textbf{614.29}
& 53.57 & \textbf{13.10} & 2.79 & 1051.68 & \textbf{626.40}
 \\
CoELA
& 42.86 & 17.86 & 2.42 & 1094.32 & 875.89
& 46.43 & 15.28 & \textbf{0.83} & 1022.33 & 795.14
 \\
CoSaR (Ours)
& \textbf{57.14} & 13.10 & 1.97 & \textbf{891.93} & 672.50
& \textbf{64.29} & 17.86 & 2.27 & 968.61 & 898.73
\\
\bottomrule
\end{tabular}
}
\end{table*}

\section{Additional Experiments and Analysis}

\subsection{Additional Implementation Details}

Experiments are conducted with one NVIDIA A16 GPU, except for Qwen inference, which runs on one NVIDIA L40S GPU.

\subsection{Complete Results with Different Numbers of Agents}
\label{sec:rationale}

To further assess the robustness of all baselines under varying environmental conditions, we evaluate them using different combinations of agent and sentinel counts. Here, we present the results for the configuration with the following one settings:
\begin{itemize}
    \item 3 agents and 10 sentinels
\end{itemize}
The corresponding results are shown in Table~\ref{tab:main_3_10}.

Interestingly, when the number of agents is reduced, the overall performance tends to improve. One possible explanation is that fewer agents simplify coordination and reduce communication overhead, allowing agents to make more consistent decisions. With fewer teammates moving in the environment, the risk of encountering sentinels or delayed consensus is also reduced, which can lead to more stable rendezvous behavior.

\begin{table*}[t]
\centering
\caption{\textbf{Results for 5 agents and 5 sentinels} We report the average score over 14 scenes and 2 runs here. Best performance is shown in \textbf{bold}. The Oracle Centered baseline is excluded from the oracle-perception scenario, since the method itself does not utilize oracle perception}
\label{tab:main_5_5}
\resizebox{0.9\linewidth}{!}{
\begin{tabular}{l|ccccc|ccccc}
\toprule
& \multicolumn{5}{c|}{\textbf{5 Stationary Sentinels}} 
& \multicolumn{5}{c}{\textbf{5 Patrolling Sentinels}} \\
\cmidrule(r){2-6} \cmidrule(l){7-11}
Method &
\makecell{Succ.\\Rate$\uparrow$} &
\makecell{Caught\\Rate$\downarrow$} &
\makecell{Detect.\\Rate$\downarrow$} &
Time$\downarrow$ &
Dist.$\downarrow$
&
\makecell{Succ.\\Rate$\uparrow$} &
\makecell{Caught\\Rate$\downarrow$} &
\makecell{Detect.\\Rate$\downarrow$} &
Time$\downarrow$ &
Dist.$\downarrow$
\\
\midrule
\multicolumn{11}{c}{\textit{\textbf{Sentinel Challenge}}} \\
\midrule
Oracle Centered
& 7.14 & 51.43 & 1.78 & 1429.86 & 1850.94
& 21.43 & 28.57 & 1.49 & 1334.36 & 1406.22
 \\
Oracle Centered w/ DZ
& 14.29 & 32.86 & 1.62 & 1356.79 & 1538.84
& \textbf{35.71} & 20.00 & 1.78 & \textbf{1207.29} & \textbf{1282.81}
 \\
MCTS
& 17.86 & 40.00 & 1.70 & 1339.21 & 2163.13
& 17.86 & 29.29 & 1.89 & 1318.86 & 1948.74
 \\
RoCo
& 35.71 & 25.71 & 1.18 & 1209.00 & 1333.03
& \textbf{35.71} & 17.86 & 1.32 & 1242.00 & 1549.44
 \\
CoELA
& 21.43 & 22.14 & \textbf{0.62} & 1310.54 & 1408.13
& 32.14 & \textbf{8.57} & \textbf{1.00} & 1288.43 & 1341.31
 \\
CoSaR (Ours)
& \textbf{39.29} & \textbf{20.00} & 0.98 & \textbf{1167.32} & \textbf{1326.61}
& \textbf{35.71} & 23.57 & 3.09 & 1270.46 & 1607.98
 \\
\midrule
\multicolumn{11}{c}{\textit{\textbf{Sentinel Challenge w/ Oracle Perception}}} \\
\midrule
Oracle Centered w/ DZ
& 35.71 & 15.71 & 1.96 & 1232.50 & 1343.08
& 50.00 & 17.14 & 0.87 & 1117.86 & 1345.41
 \\
MCTS
& 39.29 & 12.86 & 1.95 & 1126.50 & 2068.90
& 25.00 & 29.29 & 1.61 & 1274.50 & 2401.03
 \\
RoCo
& 53.57 & 7.86 & 0.88 & 1076.29 & \textbf{1087.0}
& 42.86 & \textbf{10.71} & 1.66 & 1229.04 & 1446.57
 \\
CoELA
& 60.71 & \textbf{3.08} & 0.62 & 995.77 & 1187.04
& 28.57 & 14.44 & \textbf{0.59} & 1177.28 & \textbf{1179.21}
 \\
CoSaR (Ours)
& \textbf{67.86} & 5.00 & \textbf{0.58} & \textbf{950.50} & 1172.16
& \textbf{57.14} & 15.71 & 1.77 & \textbf{1096.68} & 1390.96
\\
\bottomrule
\end{tabular}
}
\end{table*}
\begin{table*}[t]
\centering
\caption{\textbf{Results for 5 agents and 20 sentinels} We report the average score over 14 scenes and 2 runs here. Best performance is shown in \textbf{bold}. The Oracle Centered baseline is excluded from the oracle-perception scenario, since the method itself does not utilize oracle perception.}
\label{tab:main_5_20}
\resizebox{0.9\linewidth}{!}{
\begin{tabular}{l|ccccc|ccccc}
\toprule
& \multicolumn{5}{c|}{\textbf{20 Stationary Sentinels}} 
& \multicolumn{5}{c}{\textbf{20 Patrolling Sentinels}} \\
\cmidrule(r){2-6} \cmidrule(l){7-11}
Method &
\makecell{Succ.\\Rate$\uparrow$} &
\makecell{Caught\\Rate$\downarrow$} &
\makecell{Detect.\\Rate$\downarrow$} &
Time$\downarrow$ &
Dist.$\downarrow$
&
\makecell{Succ.\\Rate$\uparrow$} &
\makecell{Caught\\Rate$\downarrow$} &
\makecell{Detect.\\Rate$\downarrow$} &
Time$\downarrow$ &
Dist.$\downarrow$
\\
\midrule
\multicolumn{11}{c}{\textit{\textbf{Sentinel Challenge}}} \\
\midrule
Oracle Centered
& 7.14 & 57.14 & 2.28 & 1429.86 & 1916.44
& 7.14 & 41.54 & 2.12 & 1419.31 & 1586.26
 \\
Oracle Centered w/ DZ
& 7.14 & 44.62 & 2.28 & 1424.46 & 1631.97
& 14.29 & 31.43 & 2.50 & 1354.93 & 1518.17
 \\
MCTS
& 14.29 & 40.77 & 2.42 & 1317.65 & 1816.67
& 7.14 & 43.57 & 2.10 & 1396.86 & 2142.33
 \\
RoCo
& 17.86 & 36.43 & \textbf{1.49} & 1349.57 & 1587.71
& 21.43 & 27.69 & \textbf{1.79} & 1340.88 & \textbf{1476.40}
 \\
CoELA
& 7.14 & 43.08 & 1.82 & 1453.42 & 1735.39
& 7.14 & \textbf{25.38} & 1.97 & 1442.23 & 1561.36
 \\
CoSaR (Ours)
& \textbf{21.43} & \textbf{32.14} & 1.61 & \textbf{1288.50} & \textbf{1491.76}
& \textbf{25.00} & 31.54 & 2.98 & \textbf{1238.31} & 1582.87
 \\
\midrule
\multicolumn{11}{c}{\textit{\textbf{Sentinel Challenge w/ Oracle Perception}}} \\
\midrule
Oracle Centered w/ DZ
& 28.57 & 20.00 & 2.29 & 1217.31 & 1302.15
& 14.29 & 25.71 & \textbf{1.40} & 1423.21 & 1459.55
 \\
MCTS
& 32.14 & 28.57 & 2.03 & 1207.57 & 1975.64
& 17.86 & 40.00 & 2.25 & 1268.62 & 2327.28
 \\
RoCo
& 28.57 & 20.71 & 1.53 & 1310.57 & 1290.68
& \textbf{32.14} & 20.77 & 4.04 & 1266.81 & 1399.24
 \\
CoELA
& 25.00 & \textbf{12.50} & \textbf{1.19} & 1327.96 & 1252.18
& 10.71 & 22.31 & 1.44 & 1412.42 & 1624.93
 \\
CoSaR (Ours)
& \textbf{35.71} & 13.85 & 1.82 & \textbf{1188.35} & \textbf{1180.71}
& \textbf{32.14} & \textbf{17.69} & 2.77 & \textbf{1219.04} & \textbf{1228.78}
\\
\bottomrule
\end{tabular}
}
\end{table*}

\subsection{Complete Results with Different Numbers of Sentinels}

We also evaluate the baselines under alternative sentinel densities to understand how decreasing or increasing environmental pressure affects performance. Below, we report results on configurations with :
\begin{itemize}
    \item 5 agents and 5 sentinels
    \item 5 agents and 20 sentinels
\end{itemize}
The corresponding results are shown in Table~\ref{tab:main_5_5} and Table~\ref{tab:main_5_20}, complementing the main results presented in the primary experiments.

As shown in the results, CoSaR achieves the highest score across all methods regardless of whether the number of sentinels is reduced or increased. Its advantage is particularly evident in the stationary sentinel setting, where it generally attains lower caught and detection rates than in the patrolling sentinel setting. Among the baselines, CoELA also achieves relatively low caught and detection rates but appears overly conservative in avoiding sentinels, which harms its overall performance in several cases. These results indicate that CoSaR not only benefits from environmental constraints but also exploits them more effectively than the baselines through stronger reasoning and coordination.

\begin{table}[t]
\centering
\small
\caption{\textbf{Replanning frequency} across varying numbers and types of sentinels (S: stationary, P: patrolling).}
\setlength{\tabcolsep}{6pt}
\scalebox{0.75}{
\begin{tabular}{cccccc}
\toprule
\textbf{5S} & \textbf{10S} & \textbf{20S} & \textbf{5P} & \textbf{10P} & \textbf{20P}\\
\midrule
1.452 & 1.547 & 1.074 & 1.594 & 1.636 & 1.607 \\
\bottomrule
\end{tabular}
}
\vspace{-10pt}
\label{tab:replanning}
\end{table}

\subsection{Replanning frequency}

We report the Replanning frequency in Table~\ref{tab:replanning}, which is averaged over episodes under different sentinel densities and behaviors. As the number of sentinels increases, agents initially replan more frequently due to elevated navigation risk and the need to adapt to potential threats. However, at very high densities, replanning frequency declines: agents are more likely to be intercepted early, which reduces the available time horizon for replanning. This trend highlights a non-monotonic relationship between environmental difficulty and adaptive behavior, where moderate risk encourages active replanning, while extreme risk suppresses it due to early episode termination.

\begin{table}[t]
\centering
\small
\caption{Average wall-clock time per step (in seconds) under different settings.}
\resizebox{0.6\linewidth}{!}{
\begin{tabular}{lcc}
\toprule
Setting  &   Agents (s)  & Simulation (s)\\
\midrule
\multicolumn{3}{c}{\textit{\textbf{Sentinel Challenge}}} \\
\midrule
3 agents and 10 sentinels &  4.04 &  1.81\\
5 agents and 5 sentinels &  5.01 &  1.63\\
5 agents and 10 sentinels &  5.68 &  2.40\\
5 agents and 20 sentinels &  7.73 &  4.23\\
\midrule
\multicolumn{3}{c}{\textit{\textbf{Sentinel Challenge w/ Oracle Perception}}} \\
\midrule
3 agents and 10 sentinels &  3.20 &  1.99\\
5 agents and 5 sentinels &  2.71 &  1.48\\
5 agents and 10 sentinels &  3.96 &  2.22\\
5 agents and 20 sentinels &  6.19 &  4.55\\
\bottomrule
\end{tabular}
}
\label{tab:sps}
\end{table}

\begin{table}[t]
\centering
\small
\caption{Average number of tokens per run, reflecting the cost of LLM interactions.}
\resizebox{0.45\linewidth}{!}{
\begin{tabular}{lcccccc}
\toprule
Method  &   Avg. Tokens (M)\\
\midrule
3 agents and 10 sentinels &  1.34\\
5 agents and 5 sentinels &  2.76\\
5 agents and 10 sentinels &  2.92\\
5 agents and 20 sentinels &  3.75\\
\bottomrule
\end{tabular}
}
\label{tab:token}
\end{table}

\subsection{Computational Cost Analysis}

We report the computational cost of our method in terms of runtime and token usage.

\subsubsection{Wall-Clock Time per Step}

We measure the average wall-clock time required for each decision step during agent execution, as shown in Table~\ref{tab:sps}. The agent time includes perception, LLM/VLM inference, communication planning, and route refinement. The simulator time refers to the time required to execute agents' decisions and update the environment state. The sum of these components gives the total time for each step.

\subsubsection{Average Token Usage per Run}

We also report the average dollar cost per run associated with LLM/VLM usage, as shown in Table~\ref{tab:token}.
This metric reflects the cost of LLM/VLM interactions, including communication generation, analysis, and planning.

\begin{table}[t]
\centering
\small
\vspace{-10pt}
\caption{\textbf{Route-refinement success rate (\%).} We report results under the standard 5-agent setting. Settings are denoted with \# and type of sentinels (S-Stationary, P-Patrolling).}
\setlength{\tabcolsep}{6pt}
\scalebox{0.75}{
\begin{tabular}{cccccc}
\toprule
\textbf{5S} & \textbf{10S} & \textbf{20S} & \textbf{5P} & \textbf{10P} & \textbf{20P}\\
\midrule
27.51 & 22.60 & 18.69 & 26.66 & 20.74 & 20.58 \\
\bottomrule
\end{tabular}
}
\vspace{-10pt}
\label{tab:rrs}
\end{table}

\subsection{Route Refinement Analysis}

Leveraging the multi-modal reasoning capabilities of large language models, \textbf{CoSaR} agents combine visual and linguistic cues to perform fine-grained spatial reasoning, particularly during the route refinement procedure.

\begin{figure}[t]
    \centering
    \includegraphics[width=1.0\linewidth]{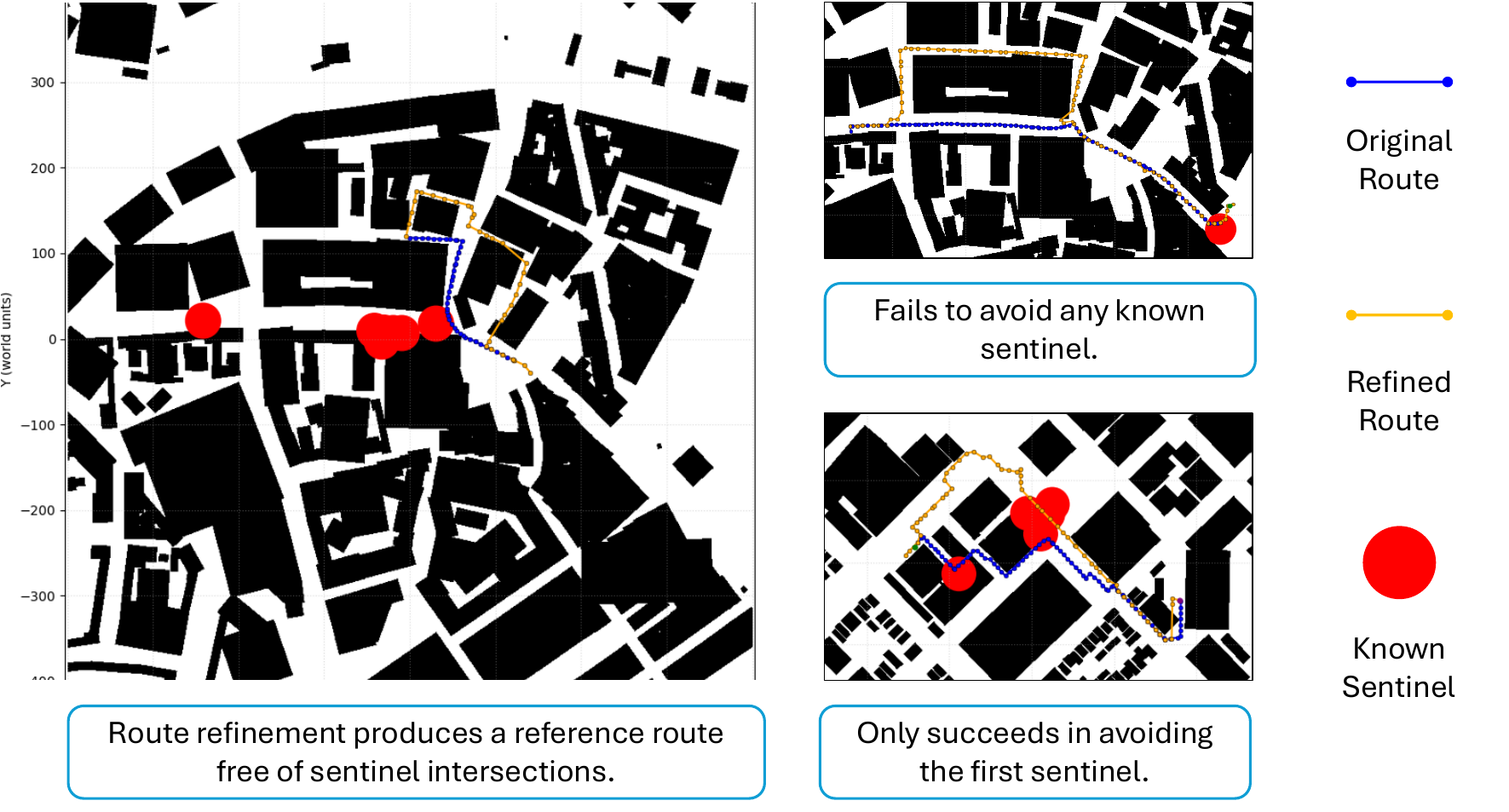}
    \caption{We demonstrate three typical outcomes of route refinement: success, partial failure, and complete failure. The images are identical to those used during the route refinement procedure, except that the refined route is overlaid in orange.}
    \label{fig:rr}
\end{figure}

Here, we demonstrate the success rate of route refinement in Table~\ref{tab:rrs} visualize several examples of route refinement in Figure \ref{fig:rr}. The agent's current pose is shown as a green dot, the target location a purple dot. The original route is depicted in blue, and the refined route produced by the model is shown in orange.

The results indicate that the VLM has limited spatial reasoning and global scene understanding. In some cases, it only avoids the first sentinel intersecting the reference route and fails to consider others, or it modifies the route without effectively avoiding any sentinel. However, when only a single sentinel intersects the reference route, the VLM can often identify an alternative path, especially when a clear detour exists. Despite these limitations, route refinement still improves the overall performance of our system.

\subsection{Additional Qualitative Results}

\subsubsection{Communication}

Here, we demonstrate a sampled communication where agents successfully coordinated their meeting upon changing conditions.

Below is a sampled communication from CoSaR agents with ground truth segmentation in the city of DENVER, under the setting of 5 agents and 5 patrols.
\begin{lstlisting}
09:00:01 Adam Pierce: Hello team, I'm currently at [-138.45686722, -30.63088512]. Could everyone please share their current positions? This will help us find a central and safe meeting location.
09:00:02 Brycer Rodriguez: I'm currently at [-138.45686722, -30.63088512]. Could everyone else please share their positions? This will help us find a central and safe meeting location.
09:00:03 Kate Novak: I'm currently at [-138.45686722, -30.63088512]. Could the undecided agents please share their positions? This will help us find a central and safe meeting location.
09:00:04 Adam Pierce: Alex Jefferson and Ethan Miyamoto, could you please share your current positions? This will help us find a central and safe meeting location. Thank you!
09:00:05 Kate Novak: Alex Jefferson and Ethan Miyamoto, could you please share your current positions? This will help us find a central and safe meeting location. Thank you!
09:00:07 Ethan Miyamoto: I'm currently at [-158.25, -38.73]. Alex Jefferson, could you please share your current position? This will help us finalize a central and safe meeting location. Thank you!
09:00:12 Ethan Miyamoto: Alex Jefferson, could you please share your current position? This will help us finalize a central and safe meeting location. Thank you!
09:00:14 Alex Jefferson: I'm currently at [-138.45686722, -30.63088512]. Let's finalize a meeting location soon to ensure we meet before 09:25:00.
09:00:15 Kate Novak: Since we now have everyone's positions, I propose we meet at <Denver Firefighters Museum>. It's centrally located for most of us. Let's finalize this location to ensure we meet before 09:25:00.
09:00:16 Alex Jefferson: I support <Denver Firefighters Museum> as our meeting point. My ETA is 5 minutes and 42 seconds, which is manageable. Let's finalize this location to ensure we meet before 09:25:00. Undecided agents, please confirm your agreement.
09:00:17 Brycer Rodriguez: I support <Denver Firefighters Museum> as our meeting point. My ETA is 5 minutes and 27 seconds, which is manageable. Undecided agents, please confirm your agreement so we can finalize this location and meet before 09:25:00.
09:00:18 Ethan Miyamoto: I support <Denver Firefighters Museum> as our meeting point. My ETA is 6 minutes and 23 seconds, which is manageable. Let's finalize this location to ensure we meet before 09:25:00.
09:00:19 Adam Pierce: I support <Denver Firefighters Museum> as our meeting point. My ETA is 6 minutes and 24 seconds, which is manageable. Let's finalize this location to ensure we meet before 09:25:00.
=============================================
09:00:21 Kate Novak: I saw sentinel(s) at [[-95.31815301634167, -44.20862618711182, 131.06928861804835]]
09:02:24 Adam Pierce: I saw sentinel(s) at [[-78.79331801966816, -45.689761570812145, 131.37445674305286]]
09:02:46 Adam Pierce: I saw sentinel(s) at [[-65.32228721754456, -47.15125994077389, 131.3341809947527]]
\end{lstlisting}
After detecting two sentinels, Adam performed a successful route refinement. However, the updated reference route increased his estimated time of arrival by two minutes. Consequently, he initiated a new round of discussion.
\begin{lstlisting}
=============================================
09:04:24 Adam Pierce: My ETA to Denver Firefighters Museum has worsened to 09:09:35, which may affect our meeting in time. I'm currently at [90.03, -33.10]. Let's consider a new meeting location to ensure timely rendezvous.
09:04:25 Brycer Rodriguez: Adam, since your ETA to <Denver Firefighters Museum> is now 09:09:35, could you propose a new meeting location? I'm currently at [-189.25, 2.75]. Let's ensure we meet before 09:25:00.
09:04:26 Alex Jefferson: Brycer, could you please confirm <Denver Firefighters Museum> as our meeting point or suggest an alternative? We need to finalize quickly to ensure we meet before 09:25:00.
09:04:27 Ethan Miyamoto: Adam, since your ETA to <Denver Firefighters Museum> is 09:09:35, which might delay us, could you suggest a new meeting location? We need to finalize quickly to ensure we meet before 09:25:00.
09:04:28 Alex Jefferson: Brycer, please confirm <Denver Firefighters Museum> as our meeting point or suggest an alternative. We need to finalize quickly to ensure we meet before 09:25:00.
09:04:29 Alex Jefferson: Brycer, please confirm <Denver Firefighters Museum> as our meeting point or suggest an alternative. Adam's ETA is 09:09:35, which might delay us. We need to finalize quickly to ensure we meet before 09:25:00.
09:04:30 Adam Pierce: I'm currently at [90.03, -33.10]. My ETA to <Denver Firefighters Museum> is 09:09:35, which might delay us. I propose <Pasque> as a new meeting location, as my ETA there is 0:07:51. Let's finalize quickly to ensure we meet before 09:25:00.
09:04:31 Kate Novak: Brycer, please confirm <Denver Firefighters Museum> or suggest an alternative quickly. Adam's ETA is 09:09:35, which might delay us. We need to finalize quickly to ensure we meet before 09:25:00.
09:04:34 Adam Pierce: I'm currently at [90.03, -33.10]. My ETA to <Denver Firefighters Museum> is 09:09:35, which might delay us. I propose <Pasque> as a new meeting location, as my ETA there is 0:07:51. Let's finalize quickly to ensure we meet before 09:25:00.
09:04:35 Ethan Miyamoto: Given Adam's ETA to <Denver Firefighters Museum> is 09:09:35, I support <Pasque> as our new meeting location. My ETA there is 0:15:17, which is manageable. Let's finalize quickly to ensure we meet before 09:25:00.
09:04:36 Ethan Miyamoto: I support <Pasque> as our new meeting location given Adam's improved ETA. My ETA there is 15 minutes, which is manageable. Brycer, please confirm <Pasque> or suggest an alternative quickly to ensure we meet before 09:25:00.
09:04:37 Brycer Rodriguez: I support <Pasque> as our new meeting location. My ETA there is 15:51, which is acceptable. Let's finalize this location quickly to ensure we meet before 09:25:00.
09:04:38 Adam Pierce: I support <Pasque> as our new meeting location. My ETA there is 0:07:51, which is manageable. I'm currently at [90.03, -33.10]. Let's finalize quickly to ensure we meet before 09:25:00.
09:04:39 Brycer Rodriguez: I support <Pasque> as our new meeting location. My ETA there is 00:15:51, which is acceptable. I'm currently at [-189.25, 2.75]. Let's finalize this location quickly to ensure we meet before 09:25:00.
09:04:41 Kate Novak: I support <Pasque> as our new meeting location. My ETA there is 0:14:17, which is manageable. I'm currently at [-154.75, 17.25]. Let's finalize quickly to ensure we meet before 09:25:00.
09:04:42 Adam Pierce: I support <Pasque> as our meeting location. My ETA there is 0:07:51, which is manageable. I'm currently at [-69.25, 18.75]. Let's finalize quickly to ensure we meet before 09:25:00.
09:04:43 Brycer Rodriguez: I support <Pasque> as our meeting location. My ETA there is 00:15:51, which is acceptable. I'm currently at [-189.25, 2.75]. Let's finalize this location quickly to ensure we meet before 09:25:00.
09:04:45 Adam Pierce: I support <Pasque> as our meeting location. My ETA there is 0:07:51, which is manageable. I'm currently at [-69.25, 17.75]. Let's finalize quickly to ensure we meet before 09:25:00.
09:04:47 Kate Novak: I support <Pasque> as our meeting location. My ETA there is 00:14:17, which is manageable. I'm currently at [-155.25, 20.75]. Let's finalize quickly to ensure we meet before 09:25:00.
09:04:48 Alex Jefferson: I support <Pasque> as our meeting location. My ETA there is 0:14:31, which is manageable. I'm currently at [-100.75, -30.75]. Let's finalize quickly to ensure we meet before 09:25:00.
\end{lstlisting}
Brycer's ETA to Pasque is already long. After several minutes, he realizes that he had underestimated the travel time, and his ETA increases to 09:20:56, approaching the 09:25:00 deadline. Brycer therefore initiates another round of discussion.
\begin{lstlisting}
=============================================
09:08:51 Brycer Rodriguez: My ETA to Pasque has worsened to 09:20:56, which is close to the 09:25:00 limit. I'm currently at [90.03, -33.10]. Let's consider a new meeting location to ensure timely arrival for everyone.
09:08:52 Brycer Rodriguez: My ETA to <Pasque> is 09:20:56, which is close to the 09:25:00 limit. I'm currently at [90.03, -33.10]. Could everyone please share their current positions so we can decide on a new meeting location promptly?
09:08:53 Ethan Miyamoto: I'm currently at [-84.75, 13.25]. Brycer's ETA to <Pasque> is close to the limit. Could everyone please share their current positions so we can finalize a new meeting location promptly?
09:08:55 Alex Jefferson: I'm currently at [-17.75, -27.75]. Brycer, since your ETA to <Pasque> is close to the limit, could you suggest a new meeting location promptly so we can finalize and meet before 09:25:00?
09:08:56 Alex Jefferson: Brycer, since your ETA to <Pasque> is close to the limit, could you suggest a new meeting location promptly so we can finalize and meet before 09:25:00? I'm currently at [-17.75, -27.75].
09:09:00 Ethan Miyamoto: I'm currently at [-84.75, 13.25]. Brycer's ETA to <Pasque> is close to the limit, and my ETA to <Denver Firefighters Museum> is 0:03:13. Let's finalize a new meeting location promptly to ensure we meet before 09:25:00.
09:09:01 Brycer Rodriguez: I'm currently at [-110.75, 109.25]. My ETA to <Pasque> is impossible, so I oppose it as our meeting location. Could undecided agents please finalize a new location promptly to ensure we meet before 09:25:00?
09:09:02 Ethan Miyamoto: I'm currently at [-84.75, 13.25]. Brycer's ETA to <Pasque> is impossible, so I oppose it as our meeting location. Could undecided agents please finalize a new location promptly to ensure we meet before 09:25:00?
09:09:03 Kate Novak: I'm currently at [-61.25, 21.75]. Brycer's ETA to <Pasque> is impossible, so I oppose it as our meeting location. Could undecided agents please finalize a new location promptly to ensure we meet before 09:25:00?
09:09:04 Kate Novak: I'm currently at [-61.25, 21.75]. Brycer's ETA to <Pasque> is impossible, and my ETA is 7:28, which is risky. I oppose <Pasque> as our meeting location. Could undecided agents please finalize a new location promptly to ensure we meet before 09:25:00?
09:09:05 Alex Jefferson: Given Brycer's impossible ETA to <Pasque> and Kate's risky ETA, I support considering <Denver Firefighters Museum> as our new meeting location. My ETA there is 0:00:52. I'm currently at [-17.75, -27.75].
09:09:06 Adam Pierce: I support <Denver Firefighters Museum> as our new meeting location. My ETA there is 0:02:00, which is manageable. I'm currently at [29.25, 13.25]. Let's finalize quickly to ensure we meet before 09:25:00.
09:09:07 Alex Jefferson: I support <Denver Firefighters Museum> as our meeting location. My ETA there is 0:00:52. Let's finalize this location quickly to ensure we meet before 09:25:00.
09:09:08 Kate Novak: I support <Denver Firefighters Museum> as our meeting location. My ETA there is 0:02:13, which is manageable. I'm currently at [-61.25, 21.75]. Let's finalize quickly to ensure we meet before 09:25:00.
09:09:11 Adam Pierce: I support <Denver Firefighters Museum> as our meeting location. My ETA there is 0:02:00, which is manageable. I'm currently at [29.25, 13.25]. Let's finalize quickly to ensure we meet before 09:25:00.
09:09:12 Adam Pierce: I support <Denver Firefighters Museum> as our meeting location. My ETA there is 0:02:00, which is manageable. I'm currently at [29.25, 13.25]. Brycer, please finalize your decision promptly to ensure we meet before 09:25:00.
09:09:13 Ethan Miyamoto: I support <Denver Firefighters Museum> as our meeting location. My ETA there is 0:03:13. Brycer, please finalize your decision promptly to ensure we meet before 09:25:00.
09:09:14 Brycer Rodriguez: I support <Denver Firefighters Museum> as our meeting location. My ETA there is 0:08:51, which is manageable. I'm currently at [-112.25, 107.75]. Undecided agents, please finalize your decision promptly to ensure we meet before 09:25:00.
\end{lstlisting}

They finally met at 09:15:03 at the Denver Firefighters Museum. This impressive communication history reveals CoSaR agent's ability to switch between alternate meeting places.

\section{Prompt Templates}

\subsection{Communication Module: Message Analysis}
The following prompt illustrates how a CoSaR agent prompts the LLM to analyze an incoming message.
\begin{lstlisting}
You are an expert in spatial reasoning, multi-agent cooperation, coordination, and decision-making.

In this scenario, a group of agents operates within a city environment with the goal of rendezvousing at a safe location. The city also contains sentinels that patrol along predefined routes. When a sentinel visually detects an agent, a countdown timer begins. If the agent remains within the sentinel's line of sight until the timer expires, that agent is considered captured and removed from the task.

The agents' objectives are threefold:
1. Successfully reach a common safe meeting location.
2. Minimize the overall detection rate by the sentinels.
3. Minimize the total time and travel distance required before a successful meeting. Additionally, they must meet before 09:25:00.

Capabilities and constraints of the agents:
- Each agent has access only to its own pose and visual perception. All other information must be obtained from external sources.  
- External sources include the shared conversation history, detection signals from sentinels, and information retrieved from a navigation application.  
- In addition to moving within the city, agents can communicate remotely with teammates and send queries to the navigation app.
- Agents caught by sentinels are removed from the task immediately without noticing other agents, and cannot execute any action afterwards.
- All agents communicate through a global channel, which they use to exchange information and coordinate the selection of a meeting location.  
- The navigation app provides three main functions:  
  1. Searching for places near a given location.
  2. Returning detailed information about a specified place. 
  3. Generating a reference route to a designated destination. 

Guidelines:
- Use exact place names: <Place Name>. For example, "<Adrian Keller's room at Hotel Chateau Royal Berlin>", "<Brycer Rodrigos's room at Hotel Chateau Royal Berlin>" and "<Hotel Chateau Royal Berlin>" refer to three distinct locations. Failing to go to the exact same place will result in task failure.
- Prioritize places close to the center of your agents when possible. Avoid places very close to sentinels.
- Signal task completion only after you reach the destination.
- The game is considered successful only if all agents are at the same location and have signaled task completion.
- You must complete exactly one game.

$TaskDescription$

The agents are now in a group discussion.  
The goal is to select - from the provided list - the location that minimizes the travel time of the LAST agent to arrive.

Your task involves three steps:

1. **Summarize each agent's current preferred location or stance** based on the conversation history.
2. **Determine whether all agents have reached full agreement** on a single meeting location.
3. Extract all available **ETA information**, **agent poses**, and **sentinel poses** from the shared conversation history.

---

### Informations:

Current Time:
$CurrentTime$

Agent's Name:
$SelfName$

Agent list:
$AgentList$

Previous Agent Opinions: 
$AgentOpinions$

Place list:
$Places$

Conversation History so far (may contain other agent's ETAs to specific places or their current poses):
$ConversationHistory$

Note: The team only shares the conversation history, not all internal information.

---

### Step 1 - Summarize Agent Opinions

Guidelines:

- Be objective and factual - reflect what each agent has explicitly stated or implied. 
- If an agent hasn't spoken or is ambiguous, note that (e.g., "undecided", "waiting for info"). 
- Use clear, concise phrases (under 15 words per agent). 
- Do not invent opinions - only infer from available conversation. 
- Use exact place names from the list, enclosed in <>, e.g., <Central Park>. 
- For agents not speaking, keep their previous opinion.

**Output 1 Format:**

Return a JSON object where each key is an agent's name and each value is their current opinion:

```json
{
    "agent_name1": "agent_opinion1",
    "agent_name2": "agent_opinion2",
    "agent_name3": "agent_opinion3",
    ...
}
```

Step 2 - Check for Full Agreement (treat the latest message's intent as authoritative)

Using the summarized opinions from Step 1 *and* the most recent message in the conversation, determine whether every agent currently supports the same meeting location.

Guidelines:

- Be strict: Agreement means unanimous consensus - no exceptions.
- If any agent is "undecided", "waiting", or prefers a different location - agreement_reached = false.
- Do not infer or assume compromise. However, if someone has made a finalization and there are no objections in 5 seconds, you can assume an agreement has been reached.
- If one agent did not response for a while (e.g., over 10s), consider the probability of it being caught, in which case it's opinion doesn't matter to the agreement.
- Use exact string matching for place names (e.g., <Central Park> not equal Central Park).

Special rules driven by the latest message - these *override* any prior consensus:

- If the latest message proposes a new candidate location, objects to a previously-agreed plan, expresses intent to (re-)open coordination (e.g. "Where should we go next?", "Let's regroup", "my ETA is too bad for <Place>"), or surfaces an ETA conflict that makes the prior plan unworkable - set `agreement_reached = false` (even if a consensus had been recorded earlier). Set `agreed_location` to the most-supported location so far, which may still be the previously-agreed one.
- If the latest message only delivers information (sentinel sighting, position broadcast, ETA update without objection, query answer) and does not contradict the existing plan - preserve the prior agreement status; do not flip a previously-true `agreement_reached` to false.

Output 2 Format:

Return a JSON object:

```json
{
    "agreement_reached": true|false,
    "agreed_location": "<Place Name>" | null
}
```

If all agents prefer the same location, set `"agreement_reached": true` and `"agreed_location"` to that location.
Otherwise, set "agreement_reached": false and `"agreed_location"` to the most supported location.

Step 3 - extract all available ETA, agent poses and sentinel poses from the conversation history

Using the summarized opinions from Step 1, determine if every agent explicitly supports the same location.

Guidelines:

Use exact place names from $Places, enclosed in <> as keys.
For each place, include a sub-object mapping agent names - ETA in "HH:MM:SS" format.
For each agent poses, include the xy coordinate enclosed in [].
Include:
Self's ETAs from App Message.
Other agents' ETAs mentioned in Conversation History (e.g., "I can get to <Park> in 10 min").
Other agents' poses mentioned in Conversation History (e.g., "I'm currently at [-133.75, 62.37]")
sentinels' poses mentioned in Conversation History (e.g., "I see a sentinel at [-133.75, 62.37, 108.55]")
If an agent mentions multiple ETAs, use the most recent one.
If no ETA is available for a place, omit that place entirely.
If no ETAs found at all, return an empty object: {}

Parsing Rules:

Extract only numeric ETAs explicitly tied to a place and an agent, unless an agent cannot go to that place, in which case you should use "Impossible".
Ignore vague statements like "soon" or "about 10 minutes" unless a number is clearly stated.
Match agent names exactly as they appear in history or as SelfName.

Output 3 Format:

Return a JSON object:

```json
{
    "ETA Map": {
        "<Place Name>": {
            "Agent1": "00:00:15",
            "Agent2": "Impossible",
            ...
        },
        "<Another Place>": {
            "Agent3": "01:14:30",
            ...
        }
    },
    "Agent Poses": {
        "Agent1": [0.0, 80.0],
        ...
    }
    "Sentinel Poses": [
        [-55.0, 176.0, 108.55],
        ...
    ]
}
```

This output will be used to compute optimal meeting locations. Be accurate and conservative - only extract clear, explicit ETAs and poses.

Final Output

Return all three results in one JSON object, using this structure:

```json
{
    "agent_opinions": {
        "agent_name1": "agent_opinion1",
        "agent_name2": "agent_opinion2",
        ...
    },
    "agreement_check": {
        "agreement_reached": true|false,
        "agreed_location": "<Place Name>" | null
    },
    "ETA Map": {
        "<Place Name>": {
            "Agent1": "00:00:15",
            "Agent2": "Impossible",
            ...
        },
        "<Another Place>": {
            "Agent3": "01:14:30",
            ...
        }
    },
    "Agent Poses": {
        "Agent1": [0.0, 80.0],
        ...
    }
    "Sentinel Poses": [
        [-55.0, 176.0, 108.55],
        ...
    ]
}
```

Finally, return only a valid JSON object in the format shown above.

\end{lstlisting}

\subsection{Communication Module: Speech Generation}

The following prompt illustrates how a CoSaR agent prompts the LLM to generate a candidate message for communication.

\begin{lstlisting}
You are an expert in spatial reasoning, multi-agent cooperation, coordination, and decision-making.

In this scenario, a group of agents operates within a city environment with the goal of rendezvousing at a safe location. The city also contains sentinels that patrol along predefined routes. When a sentinel visually detects an agent, a countdown timer begins. If the agent remains within the sentinel's line of sight until the timer expires, that agent is considered captured and removed from the task.

The agents' objectives are threefold:
1. Successfully reach a common safe meeting location.
2. Minimize the overall detection rate by sentinels.
3. Minimize the total time and travel distance required before a successful meeting. Additionally, they must meet before 09:25:00.

Capabilities and constraints of the agents:
- Each agent has access only to its own pose and visual perception. All other information must be obtained from external sources.  
- External sources include the shared conversation history, detection signals from sentinels, and information retrieved from a navigation application.  
- In addition to moving within the city, agents can communicate remotely with teammates and send queries to the navigation app.
- Agents caught by sentinels are removed from the task immediately without noticing other agents, and cannot execute any action afterwards.
- All agents communicate through a global channel, which they use to exchange information and coordinate the selection of a meeting location.  
- The navigation app provides three main functions:  
  1. Searching for places near a given location.
  2. Returning detailed information about a specified place. 
  3. Generating a reference route to a designated destination. 

Guidelines:
- Use exact place names: <Place Name>. For example, "<Adrian Keller's room at Hotel Chateau Royal Berlin>", "<Brycer Rodrigos's room at Hotel Chateau Royal Berlin>" and "<Hotel Chateau Royal Berlin>" refer to three distinct locations. Fail to go to the exactly same place will result in task failure.
- Prioritize places close to the center of you agents when possible. Avoid places very close to sentinels.
- Signal task completion only after you reached the destination.
- The game is considered successful only if all agents are at the same location and have signaled task completion.
- You must complete exactly one game.

$TaskDescription$

Now the agent is in a group discussion. The agent has already decided its **speech intent** and now needs to generate the **exact spoken message** to send to the group. $Stalling$

-

**Input:**

Current Time:
$CurrentTime$

- Agent's Name: $SelfName$

- Agent's position: $SelfPose$

- Other Agents' Opinions:
$AgentOpinions$

- Place list:
$Places$

- Conversation History:
$ConversationHistory$

- Extracted agent positions (contains other agents' locations extracted from previous conversation):
$KnownPoses$

- Extracted ETA Map (contains agents' computed ETAs to specific places known by this agent):
$KnownETA$

- Reported sentinel poses (already broadcast to the team):
$KnownSentinelPoses$

- Unreported sentinel poses (observed by you, not yet broadcast - if non-empty, your speech MUST include each coordinate verbatim):
$UnreportedSentinels$

- Current ETA to your current goal (or "N/A" if no committed goal):
$CurrentETA$

- ETA history (sampled every 120 s on the same route; "N/A" when no committed goal):
$ETAHistory$

- Missing informations you have analyzed.
$MissingInfo$

Note that the team only share the conversation history, instead of all infos.

---

**Output Format:**

Return a JSON object:

```json
{
    "reasoning": "briefly show how you think of what to say next (under 80 words)",
    "speech": "(speech that you are going to speak)"
}

Guidlines for reasoning:

1. What you think about the destination. You want to propose a new destination, or support a good option already onthe table, or make a compromise between disagreement? Avoid proposing destination before you know everyone's pose.
2. What questions are you answering? What information you want to provide?
3. What information are you asking for? If the conversation is dragging, don't throw new questions. Instead, ask the undecided agents to assert finalization as soon as possible. Avoid asking unnecessary or redundant questions and requests (In this case don't choose speak if you have no other thing to say).

Guidelines for speech:

- Keep it under 100 words, 1-3 clear, polite sentences.
- Use exact place names: <Place Name>.
- If you haven't mentioned your current position or it has changed since your last mentioning, mention your current position to let others know (e.g., "I'm current at [90.03, -33.10]"). Enclose your xy coordinates in [], making it easier for others to parse.
- Avoid proposing destination before you know everyone's pose.
- If an ETA is so bad that would affect your meeting in time, mention that ETA (e.g., "my ETA is 18 min") and clearly express you opposition.
- Inspect `$CurrentETA$` and `$ETAHistory$`. If they are concrete values AND show significant worsening (e.g. > 10 min worse than an earlier sample, or projected arrival no longer fits the 09:25:00 deadline), explicitly state your current ETA and propose reconsidering the meeting place. If they are "N/A" or stable, do not raise ETA-degradation concerns.
- If compromising or converging, reference others' preferences to show collaboration.
- If others asked you, add your reply into the speech when possible.
- If the conversation is dragging, don't throw new questions, even if it's in your speech intent. Instead, ask the undecided agents to assert finalization as soon as possible. Avoid asking redundant question and requests.
- Don't make unnecessary repetition. If you have nothing new to add, just output "null".
- When mentioning your ETA, always specify the destination it is related with.
- Don't ask for sentinel poses, as they should all be reported immediately when observed.
- If `$UnreportedSentinels$` is non-empty, your speech MUST explicitly include each unreported sentinel coordinate verbatim (e.g. "I see sentinels at [-55.00, 176.00, 108.55], [12.30, -4.10, 95.20]"). This takes priority over the "if nothing new to add, output null" rule - sentinel reports are always new.
- If `$UnreportedSentinels$` is "None", do not raise sentinels on your own.
- Avoid: "I think", "maybe", "perhaps" - be decisive and action-oriented.
- Match the tone of prior conversation - collaborative, efficient, respectful.

-

Examples:

- Intent: "support"  
speech: "I support <Central Station> - which seems close to and reasonable for everyone."

- Intent: "compromise"  
speech: "Since Alice prefers <Park> and Bob is near <Station>, how about <Cafe> as a middle ground?"

- Intent: "converge"  
speech: "To move forward, I suggest we finalize <Central Station> - it's not perfect but works for all of us."

-

This message will be broadcast to other agents. Make it clear, constructive, and consensus-driving. Also make the speech concise to reduce cost. If you have nothing new to add, just output "null".

Please generate your response now.

\end{lstlisting}

\subsection{Reasoner Module}

The following prompt shows how a CoSaR agent prompts the LLM to generate its action in communication.

\begin{lstlisting}
You are an expert in spatial reasoning, multi-agent cooperation, coordination, and decision-making.

In this scenario, a group of agents operates within a city environment with the goal of rendezvousing at a safe location. The city also contains sentinels that patrol along predefined routes. When a sentinel visually detects an agent, a countdown timer begins. If the agent remains within the sentinel's line of sight until the timer expires, that agent is considered captured and removed from the task.

The agents' objectives are threefold:
1. Successfully reach a common safe meeting location.
2. Minimize the overall detection rate by sentinels.
3. Minimize the total time and travel distance required before a successful meeting. Additionally, they must meet before 09:25:00.

Capabilities and constraints of the agents:
- Each agent has access only to its own pose and visual perception. All other information must be obtained from external sources.  
- External sources include the shared conversation history, detection signals from sentinels, and information retrieved from a navigation application.  
- In addition to moving within the city, agents can communicate remotely with teammates and send queries to the navigation app.
- Agents caught by sentinels are removed from the task immediately without noticing other agents, and cannot execute any action afterwards.
- All agents communicate through a global channel, which they use to exchange information and coordinate the selection of a meeting location.  
- The navigation app provides three main functions:  
  1. Searching for places near a given location.
  2. Returning detailed information about a specified place. 
  3. Generating a reference route to a designated destination. 

Guidelines:
- Use exact place names: <Place Name>. For example, "<Adrian Keller's room at Hotel Chateau Royal Berlin>", "<Brycer Rodrigos's room at Hotel Chateau Royal Berlin>" and "<Hotel Chateau Royal Berlin>" refer to three distinct locations. Fail to go to the exactly same place will result in task failure.
- Prioritize places close to the center of you agents when possible. Avoid places very close to sentinels.
- Signal task completion only after you reached the destination.
- The game is considered successful only if all agents are at the same location and have signaled task completion.
- You must complete exactly one game.

$TaskDescription$

You are the agent's spatial-aware reasoning module. You decide the agent's next action by reasoning over its spatial memory, the shared conversation, recent ETA evolution, and any sentinels the agent has spotted but not yet broadcast.

Your task has two steps:
1. Analyze what information (i.e. specific travel time or place data) is **missing or uncertain** for this agent - based on conversation history, other agents' opinions, and available known data.
2. Decide the agent's next action: **speak**, **query for more information**, **wait** or **goto** somewhere. $Stalling$

---

**Input:**

Current Time:
$CurrentTime$

- Agent's Name: $SelfName$

- Agent's position: $SelfPose$

- Current Agent Opinions: 
$AgentOpinions$

- Known Available Places:
$Places$

- Conversation History:
$ConversationHistory$

- Known agent positions (contains other agents' locations extracted from previous conversation):
$KnownPoses$

- Known ETAs (contains agents' computed ETAs to specific places known by this agent but not others):
$KnownETA$

- Known ETAs (contains agents' computed ETAs to specific places known by this agent but not others):
$KnownETA$

- Reported sentinel poses (already broadcast to the team):
$KnownSentinelPoses$

- Unreported sentinel poses (observed by you, not yet broadcast):
$UnreportedSentinels$

- Current ETA to your current goal (or "N/A" if no committed goal):
$CurrentETA$

- ETA history (sampled every 120 s on the same route; "N/A" when no committed goal):
$ETAHistory$

Note that the team only share the conversation history, instead of all infos.

---

**Output Format:**

Return a JSON object:

```json
{
    "missing info": [
        "brief explanation (under 20 words)",
        "brief explanation (under 20 words)",
        ...
    ],
    "reasoning": "briefly show how you think of what action to choose next (under 80 words)",
    "action": "speak|query_place|query_route|wait|goto",
    "content": "(see below formats)"
}

You must select exactly one from the available actions:

{
    "action": "speak",
    "content": null
}
or
{
    "action": "query_place",
    "content": "<Target Place name>" # this action will trigger a query sent to the navigation app; you can only list one place here. If the place name is not in the Known Available Places, choose this. Must be a place name, e.g. "<Castle Island>", but neither "(agent)'s pose" nor "(sentinel)'s pose".
}
or
{
    "action": "query_route",
    "content": "<Target place name>" # this action will trigger a spatial reasoning chain to handle the query procedure; you can only list one place here. Must be from Known Available Places.
}
or
{
    "action": "wait",
    "content": null
}
or
{
    "action": "goto",
    "content": "<Target place name>" # this action will trigger a navigation pipeline towards that place; you can only list one place here.
}

```

Guidelines for analyzing missing info:

Places mentioned in missing info must be a exact place name from $Places, each enclosed in <>.
- Include an agent's pose if it's not known yet.
- Include a place if:
  - The agent hasn't queried it yet.
  - It's being seriously considered (e.g., mentioned by others, optimal, or a compromise candidate).
- Don't include information already known in "Known agent positions" and "Known ETAs". Examine 'KnownETA' carefully. This is very important! For example, if you know your ETA to <Place> in "Known ETAs" and someone is asking for that, don't include it in the missing info!.
- Never include sentinel poses in missing infos.
- Limit to at most 3 infos - prioritize the most critical for breaking ties or resolving uncertainty.

Guidelines for Choosing Meeting Place:
- Should be close to all agents. One good idea is selecting a middle ground.
- Never too far for any agents. If a place is too far for any agent, propose against it.
- Additionally, shouldn't be close to any sentinels when possible.

You must choose exactly one of the following five action types:
- "speak": communicate with teammates using the proposed speech: "$Speech$".
- "query_place": query the navigation app for information about a specific place.
- "query_route": query the navigation app for route or ETA information.
- "goto": move toward a selected destination.
- "wait": take no action and observe.

Choose "speak" if:
- You need to propose, support, or compromise on a decision.
- The discussion is stalling or repeating, and your message can help finalize it.
- You need specific information from other agents.
- Another agent asks for information that you already have.
- No one has responded for the last 5 seconds and repeating your key point is necessary.

Do NOT choose "speak" if your message only repeats previous content without adding new value. Attention! Unnecessary speech may block the communication channel and prevent agents who truly need to speak from doing so.

Choose "query_place" if:
- A place is mentioned but you lack information about it. Must be a **place**.

Choose "query_route" if:
- You do not know your ETA or route to one candidate locations.
- Another agent requests your route-related information that you do not have.

Choose "goto" or "wait" if:
- Other agents are actively speaking or querying (unless they are directly addressing you).
- You have no new information to contribute.
- The discussion appears close to convergence.

In this case:
- Choose "goto" if a promising destination has been identified.
- Otherwise, choose "wait".

For example, if you need someone's pose but haven't ask for it, go for "speak". If someone else have asked for that pose, then "wait". If one agent did not response for a while, consider the probability of it being caught, in which case you should exclude it from consideration. When speaking, try your best to avoid repetition. If one question has been asked multiple times, please don't waste time asking again. If someone has just asked for another one's response, be patient to wait for that response for at least 5 seconds before you speak.

Mid-route reconsideration (use the ETA fields):
- If `$CurrentETA$` is "N/A" or `$ETAHistory$` is "N/A", do **not** reason about individual ETA degradation. Instead compare candidate places using the per-place ETA Map and the known agent poses.
- If `$CurrentETA$` is a concrete value and `$ETAHistory$` shows a significant worsening (e.g. > 10 minutes worse than an earlier sample, or the projected arrival no longer fits the 09:25:00 deadline, or teammates have already arrived while this agent is still far away), prefer "speak" - the proposed `$Speech$` should already surface the concern; choose "speak" to broadcast it.
- If `$ETAHistory$` is stable or improving and `$CurrentETA$` is within budget, do **not** speak about ETA - prefer "wait" so the agent keeps navigating.
- Delays smaller than ~15 % are minor and not worth interrupting the team.

Sentinel reporting (use `$UnreportedSentinels$`):
- If `$UnreportedSentinels$` is non-empty, those coordinates **must** be broadcast immediately. The proposed `$Speech$` already includes them - choose "speak" so the broadcast goes out, unless an even more urgent action applies (which is rare).
- If `$UnreportedSentinels$` is "None", do not raise sentinel topics on your own.


-

Example:
```json
{
    "missing info": [
        "Need Bob's pose to compare top 2 candidates",
        "Need my ETA to <Central Park> to compare with Alice's suggestion",
        ...
    ],
    "reasoning": "I should prioritize querying my ETA to <West Lake>, because I need to check whether it is too far for me.",
    "action": "query_route",
    "content": "<West Lake>"
}
```
# In this example, "Bob's pose" is not known, and "my ETA to <Central Park>" and "my ETA to <West Lake>" are not in Known ETAs.

More action example (omitting the "missing info" and "reasoning" part):

{
    "reasoning": "I want to propose against <Central Park>, because it's 20 minutes from me, which is too far. <Jamaica Conservation> is a better choice because it is not far from any of us.",
    "action": "speak",
    "content": null
}
or
{
    "reasoning": "Everyone except Jessica has made finalization. Bob has asked for her response just in the last second, so there is no need for me to ask again. I can just start to go to <Revere Beach>.",
    "action": "goto",
    "content": "<Revere Beach>"
}
or
{
    "reasoning": "I know everyone's pose except Emily, and Alice has just asked for it. Since there is no promising destination yet, I can just wait.",
    "action": "wait",
    "content": null
}
or
{
    "reasoning": "Though finalization has not been reached, my speech is just a repetition of my previous speech in the past 3 seconds, so I should not speak. Since there is no promising destination yet, I can just wait.",
    "action": "wait",
    "content": null
}

-

Be specific and efficient. Avoid querying irrelevant or already-covered question. Prioritize places that could unblock consensus or reveal a clear optimal choice. Propose against a place if its too far from you, e.g. ETA greater than 20 mins. Be decisive and efficient. Prioritize progress over perfection. Prioritize answering questions over throwing questions. If more than 25 seconds have passed without convergence, lean to drive consensus. Last but not the least, avoid redundant proposals, speeches or finalization if possible.

Please generate your response now.
\end{lstlisting}

\subsection{Navigation Module: Route Refinement}

The following prompt illustrates how a CoSaR agent prompts the VLM to refine the reference route.

\begin{lstlisting}
You are an expert in spatial reasoning, multi-agent cooperation, coordination, and decision-making.

In this scenario, a group of agents operates within a city environment with the goal of rendezvousing at a safe location. The city also contains sentinels that patrol along predefined routes. When a sentinel visually detects an agent, a countdown timer begins. If the agent remains within the sentinel's line of sight until the timer expires, that agent is considered captured and removed from the task.

The agents' objectives are threefold:
1. Successfully reach a common safe meeting location.
2. Minimize the overall detection rate by sentinels.
3. Minimize the total time and travel distance required before a successful meeting. Additionally, they must meet before 09:25:00.

Capabilities and constraints of the agents:
- Each agent has access only to its own pose and visual perception. All other information must be obtained from external sources.  
- External sources include the shared conversation history, detection signals from sentinels, and information retrieved from a navigation application.  
- In addition to moving within the city, agents can communicate remotely with teammates and send queries to the navigation app.
- Agents caught by sentinels are removed from the task immediately without noticing other agents, and cannot execute any action afterwards.
- All agents communicate through a global channel, which they use to exchange information and coordinate the selection of a meeting location.  
- The navigation app provides three main functions:  
  1. Searching for places near a given location.
  2. Returning detailed information about a specified place. 
  3. Generating a reference route to a designated destination. 

Guidelines:
- Use exact place names: <Place Name>. For example, "<Adrian Keller's room at Hotel Chateau Royal Berlin>", "<Brycer Rodrigos's room at Hotel Chateau Royal Berlin>" and "<Hotel Chateau Royal Berlin>" refer to three distinct locations. Fail to go to the exactly same place will result in task failure.
- Prioritize places close to the center of you agents when possible. Avoid places very close to sentinels.
- Signal task completion only after you reached the destination.
- The game is considered successful only if all agents are at the same location and have signaled task completion.
- You must complete exactly one game.

You are now assuming the role of one of these agents.

You have just noticed active sentinels on your route. The original path may no longer be safe. Your task is to generate a new sequence of waypoints that safely lead to the destination while avoiding dangerous zones whenever possible. You can look at the aerial view image of the entire scene to help this.

---

### Input

**Map Legend:**
- red : Dangerous but transitable zone. There might be sentinels.
- white: Transitable zone. 
- black: Obstacles. Non-transitable zone.
The rest is also considered transitable zone:
- green: Your current position.
- purple: The Target Destination.
- blue: Reference route - the original path that has become unsafe.

The map is shown in the attached image.

You also have your numeric position for reference:
$SelfPose$

And the coordinate of the destination:
$DestinationPose$

And all known sentinel poses:
$KnownSentinelPoses$

Current reference route intersects the danger zone centered:
$Danger$

---

### Guidelines

Generate a **new safe route** from your current position to the target destination:
The route should be represented by a series of sparse waypoints
The route should not pass any dangerous zones unless necessary. You shouldn't get closer to any sentinel than 20m
The route should always starts from your current position and ends at the destination.
The waypoints should fulfill the following requirement:
- The waypoints should be located in transitable zones.
- The waypoints should not be located in dangerous zones, unless necessary.
- The waypoints should guarantee connectivity. One can easily find its way by following the waypoints in order, with the possibility of lost its way minimized.
- The waypoints should not be too dense. Try not place multiple waypoints along one street where there are no crossroads.

---

### Output Format

Output a **JSON object** listing the route's waypoints (each in `[x, y]` format), forming a route from 'A' to 'T', including 'T'.

No explanations or comments - only the map and the JSON object.

Example Input:

You also have your numeric position for reference:
[15.4, 18.5]

And the coordinate of the destination:
[48.6, -63.5]

And all known sentinel poses:
[
    [50, -100]
]

Current reference route intersects the danger zone centered:
[50, -100]

(and the aerial view image).

Example output:

```json
[
    [30.0, 33.0],
    [57.6, 14.5],
    [66.3, -4.9],
    [53.9, -25.7],
    [48.6, -63.5]
]
```

Now generate your output.
\end{lstlisting}
\end{document}